\documentclass[a4paper,fleqn]{cas-dc}
\usepackage{changepage}
\usepackage{color}
\makeatletter       
\DeclareRobustCommand\onedot{\futurelet\@let@token\@onedot}
\def\@onedot{\ifx\@let@token.\else.\null\fi\xspace}
\def\eg{\emph{e.g}\onedot} 
\def\ie{\emph{i.e}\onedot} 
 
\def\etc{\emph{etc}\onedot}

\def\aka{a.k.a\onedot}

\newcommand{\qr}[1]{\textcolor{black}{#1}}
\hypersetup{
    colorlinks=true,
    linkcolor=cyan,
    filecolor=cyan,      
    urlcolor=cyan,
    citecolor=cyan,
}
   
\usepackage[numbers]{natbib}
\def\tsc#1{\csdef{#1}{\textsc{\lowercase{#1}}\xspace}}
\tsc{WGM}
\tsc{QE}

\begin{document}
\let\WriteBookmarks\relax
\def\floatpagepagefraction{1}
\def\textpagefraction{.001}
     

\shortauthors{R. Qian, X. Lai and X. Li}  

\title [mode = title]{3D Object Detection for Autonomous Driving: A Survey}  

\author[1]{Rui Qian}
\ead{qiianruii@gmail.com} 

\address[1]{Key Lab of Data Engineering and Knowledge Engineering, 
Renmin University of China, Beijing 100872, China.}

\author[2]{Xin Lai}
\ead{laixin@ruc.edu.cn} 
\address[2]{School of Mathematics, Renmin University of China, Beijing 100872, China}

\author[1]{Xirong Li}
\cormark[1] 
\ead{xirong@ruc.edu.cn} 


\cortext[cor1]{Corresponding author} 

\nonumnote{
\href{https://doi.org/10.1016/j.patcog.2022.108796}{https://doi.org/10.1016/j.patcog.2022.108796}}
\nonumnote{
  \text{©} 2022 Elsevier Ltd. All rights reserved.}

\begin{abstract}
    Autonomous driving is regarded as one of the most promising remedies to shield human beings from severe 
    crashes. To this end, 3D object detection serves as the core basis of perception stack especially for the sake of 
    path planning, motion prediction, and collision avoidance \etc. \qr{Taking a quick glance at the progress we have made, 
    we attribute challenges to visual appearance recovery in the absence of depth information from images, representation 
    learning from partially occluded unstructured point clouds, and semantic alignments over heterogeneous features from 
    cross modalities.} Despite existing efforts, 3D object detection for autonomous driving is still in its infancy. 
    Recently, \qr{a large body of literature have been investigated to address this 3D vision task. Nevertheless, 
    few investigations have looked into collecting and structuring this growing knowledge. We therefore 
    aim to fill this gap in a comprehensive survey,} encompassing all the main concerns including sensors, datasets, performance
    metrics and the recent state-of-the-art detection methods, together with their pros and cons. Furthermore, 
    we provide quantitative comparisons with the state of the art. \qr{A case study on fifteen selected 
    representative methods is presented, involved with runtime analysis, error analysis, and robustness 
    analysis.} Finally, we provide concluding remarks after an in-depth analysis of the surveyed works and 
    identify promising directions for future work.   
 
    \textcolor{white}{\text{©} 2022 Elsevier Ltd. All rights reserved.}
       
    \rightline{\text{©} 2022 Elsevier Ltd. All rights reserved.}
\end{abstract}
\begin{keywords}
    \sep 3D object detection \sep Point clouds \sep Autonomous driving
\end{keywords}
   
\maketitle
\section{INTRODUCTION} \label{sec:intro}

\begin{figure}[pos=htbp] 
        \setlength{\abovecaptionskip}{0.1cm}
        \setlength{\belowcaptionskip}{0cm}
                \vspace{1.2em}
                \centering  
                  \includegraphics[width=0.49\textwidth]{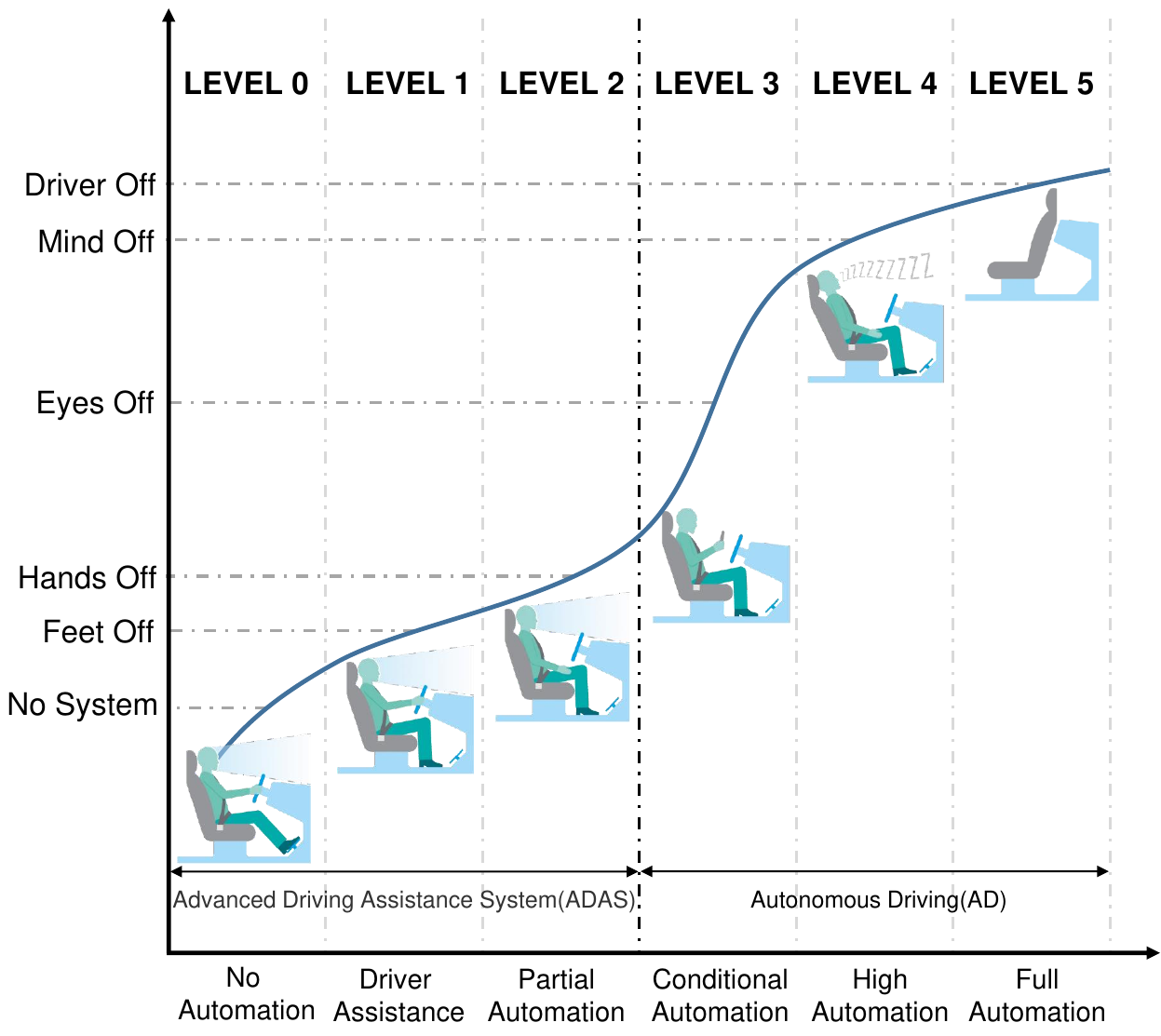} 
                \caption{\textbf{Levels of autonomous driving proposed by SAE (Society of Automotive Engineers) International \cite{sae-level}.} Where are we now?}
                \label{subfig:level5} 
             \vspace{-1em}   
\end{figure}

Dream sheds light on reality. It is a dream that autonomous vehicles hit the roads legally, 
functioning wisely as good as human drivers or even better, responding timely to various unconstrained driving 
scenarios, and being fully free of the control of human drivers, \aka Level 5 wit ``driver off'' in Fig. \ref{subfig:level5}.
Let the dream be realized, thousands of new employment opportunities shall be created for those physically 
impaired (\emph{Mobility}), millions of lives shall be rescued from motor vehicle-related crashes (\emph{Safety}), and 
billions of dollars shall be saved from disentangling traffic accidents and treating the wounded (\emph{Economics}). 
\qr{It is a reality that there is still no universal consensus on where we are now and how we shall go next. 
As illustrated in Fig. \ref{subfig:level5}, We are largely above level 2 but under or infinitely close to level 3 
by taking into account the following three social concerns: (1) Safety and Security. Rules and regulations are 
still blank, which shall be developed by governments to guarantee the safety for an entire trip. (2) Law and 
Liability. How to define the major responsibility and who will take that responsibility shall be identified 
both ethically and clearly. (3) Acceptance. Long-term efforts shall be made to establish the confidence and 
trust for the whole society, before autonomous driving can be finally accepted. This survey paper will take a 
structured glance at 3D object detection, one of the core techniques for autonomous driving.
}

\qr{      
Perception in 3D space is a prerequisite in autonomous driving. A fully understanding of what is happening 
right now in front of the vehicle will facilitate downstream components to react accordingly, which is 
exactly what 3D object detection aims for. 3D object detection perceives and describes what surrounds us via 
assigning a label, how its shape looks like via drawing a bounding box, and how far away it is from an ego vehicle via 
giving a coordinate. Besides, 3D detection even provides a heading angle that indicates orientation. It is these 
upstream information from perception stack that enables downstream planning model to make decisions.
}  
\begin{figure*}[pos=htbp]
    \setlength{\abovecaptionskip}{0.1cm}
    \setlength{\belowcaptionskip}{0cm}
        \centering     
            \includegraphics[width=0.92\textwidth]{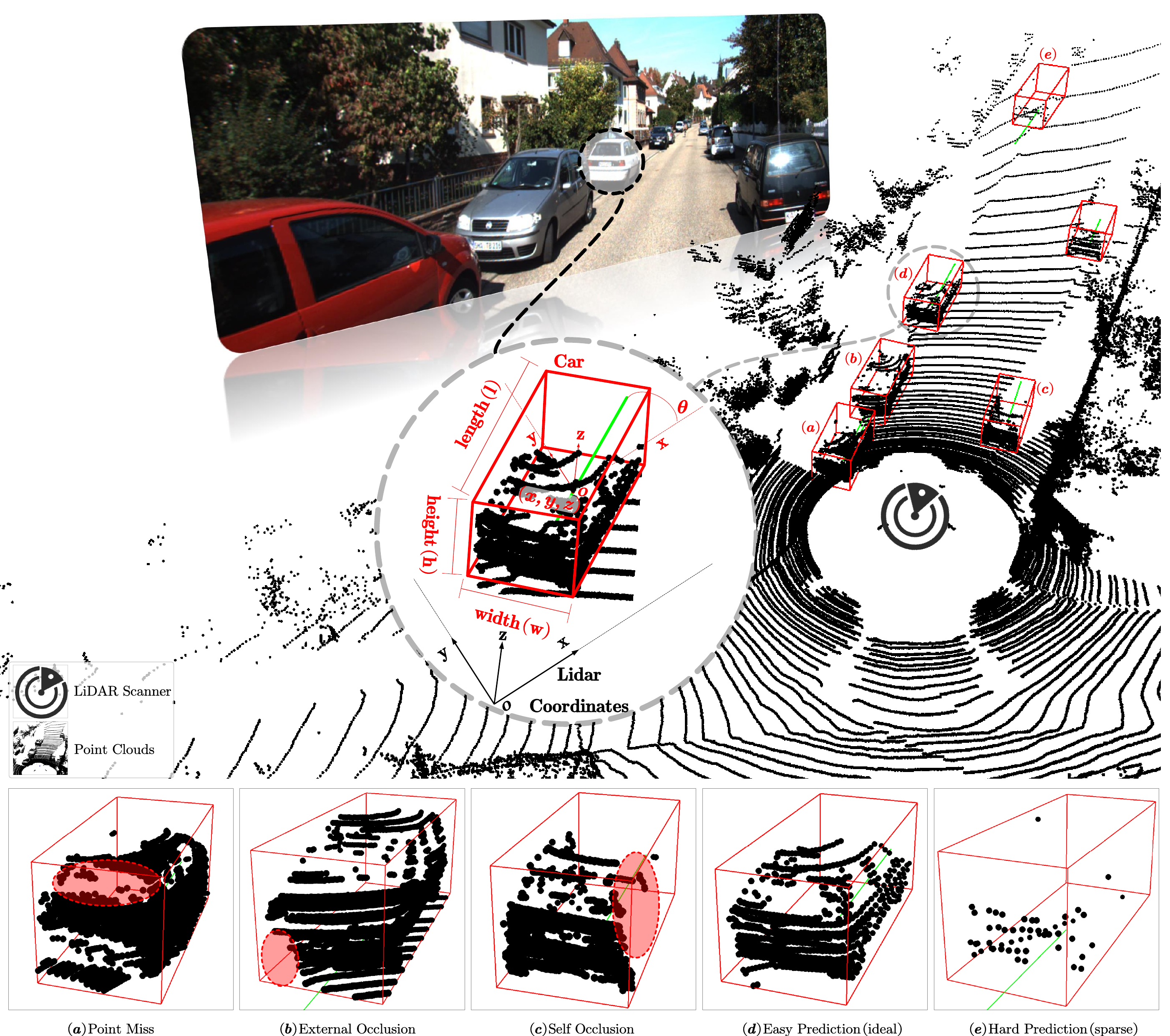} 
            \caption{\textbf{An overview of 3D object detection task from images and point clouds.}
             Typical challenges:
            (a) Point Miss. When LiDAR signals fail to return back from the surface of objects.
            (b) External Occlusion. When LiDAR signals are blocked by occluders in the vicinity. 
            (c) Self Occlusion. When one near side of the object blocks the other, which makes 
            point clouds 2.5D in practice. Note that bounding box prediction in (d) is much easier 
            than that in (e) due to the sparsity of point clouds at long ranges.}
            \label{subfig:overview}   
      \vspace{-1.5em}
\end{figure*}  

\subsection{Tasks and Challenges}
\qr{Fig. \ref{subfig:overview} presents an overview of 3D object detection task from images and point clouds.}
The whole goal of 3D object detection is to recognize the objects of interest by drawing an oriented 3D bounding 
box and assigning a label. \qr{Consider two commonly used 3D object detection modalities, \ie images and point clouds, 
the key challenges of this vision task are strongly tied to the way we use, the way we represent, and the way we 
combine. With only images on hand, the core challenge arises from the absence of depth information. Although much 
progress has been made to recover depth from images \cite{psmnet2018, dorn2019fu}, it is still an ill-posed 
inverse problem. The same object in different 3D poses can result in different visual appearance in the image plane, 
which is not conducive to the learning of representation \cite{Chen0SJ20}. Besides, given that camera is passive 
sensor (see Sec. \ref{sec:PassiveSensors}), images are naturally vulnerable to illumination (\eg, nighttime) or rainy 
weather conditions. With only point clouds on hand, the key difficulty stems from the representation learning. 
Point clouds are sparse by nature, \eg in works \cite{Mao_2021_ICCV, lang2019pointpillars}, non-empty voxels 
normally account for approximately 1\%, 3\% in a typical range setup on Waymo Open \cite{Sun_2020_CVPR} dataset 
and KITTI \cite{geiger2012we} dataset respectively. Point clouds are irregular and unordered by nature. Directly 
applying convolution operator to point clouds will incur ``desertion of shape information and variance to point 
ordering'' \cite{li2018pointcnn}. Besides, point clouds are 2.5D in practice as they are point miss in (a), 
external-occlusion in (b), self-occlusion in (c) \cite{xu2020behind}, as indicated in Fig. \ref{subfig:overview}. 
Without the umbrella of convolutional neural networks, one way is to present point clouds as voxels. The dilemma is 
that scaling up voxel size will loss resolution and consequently degrade localization accuracy while scaling down 
its size will cubically increase the complexity and memory footprints as the input resolution grows. Another way 
is to present point clouds as point sets. Nevertheless, around 80\% of runtime is occupied by point retrieval, 
say, ball query operation, in light of the poor memory locality \cite{pvconv2019}. With both images and point clouds on hand, the 
tricky obstacle often derives from semantic alignments. Images and point clouds are two heterogeneous media, presented 
in camera view and real 3D view, finding point-wise correspondences between LiDAR points and image pixels results in 
``feature blurring'' \cite{PointPainting2020Vora}.
}  
  
\begin{figure*}[pos=htbp]
    \setlength{\abovecaptionskip}{0.1cm}
    \setlength{\belowcaptionskip}{0cm}
        \centering     
            \includegraphics[width=0.955\textwidth]{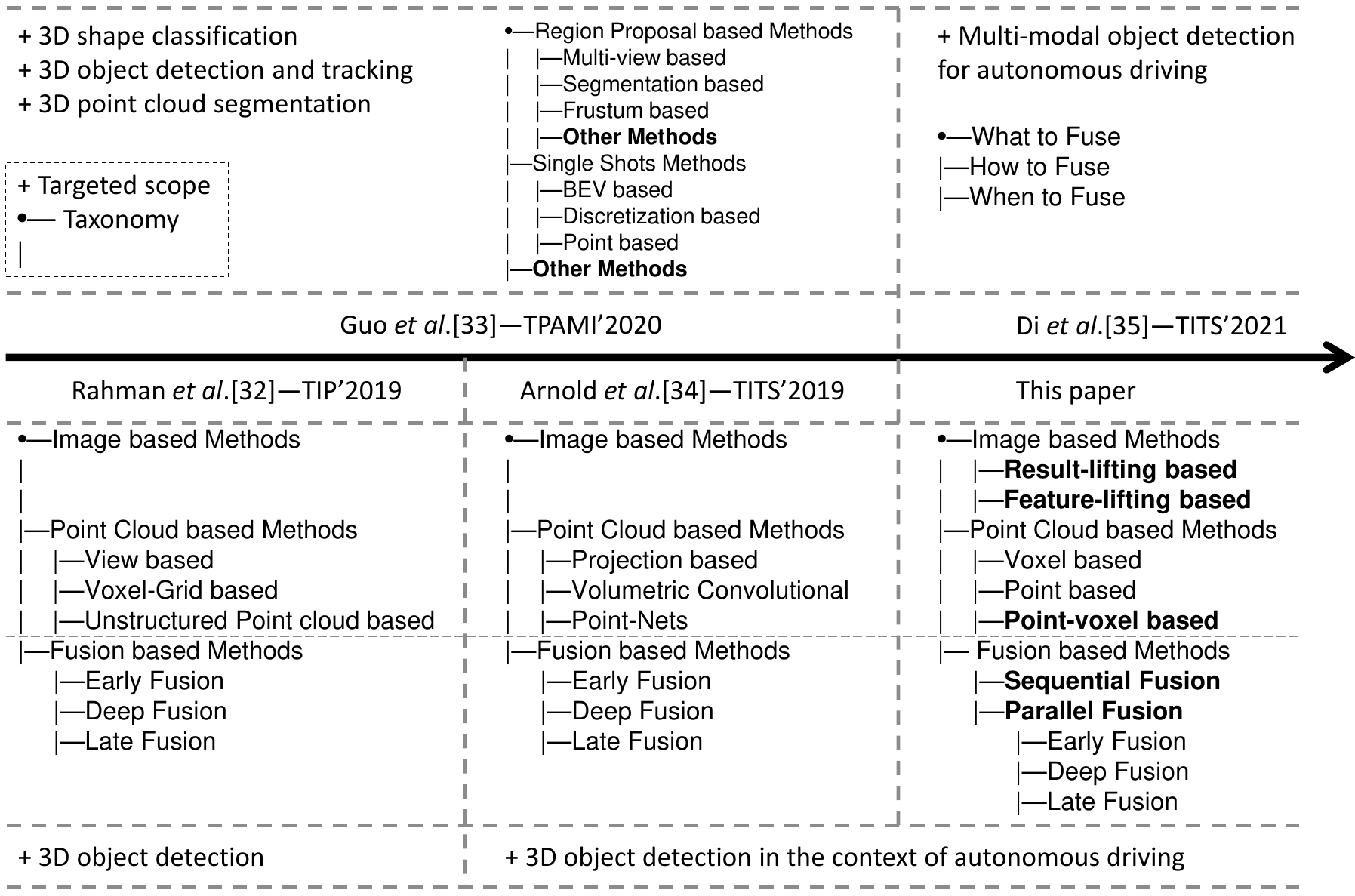} 
            \caption{\textbf{A summary showing how this survey differs from existing ones on 3D object detection.}
            Vertically, targeted scope concisely determines where the boundary is located among their investigations.
            Horizontally, hierarchical branches of this paper reveal a good continuity of existing efforts 
            \cite{rahman2019recent, arnold2019survey} while adapt new branches (indicated in bold font) for 
            dynamics, which importantly contributes to the maturity of the taxonomy on 3D object detection.
            }
            \label{subfig:comparision} 
      \vspace{-1em}   
\end{figure*} 
\subsection{Targeted Scope, Aims, and Organization}
\qr{   
We review literature that are closely related to 3D object detection in the context of autonomous driving. Depending on 
what modality we use, existing efforts are divided into the following three subdivisions: (1) image based \cite{ChabotCRTC17DeepMANTA, 
MousavianAFK17Deep3DBox, li2019gs3d, li2019stereo, Shi_2021_ICCV, Park_2021_ICCV, Ma_2021_CVPR, 2021YOLOStereo3D}, 
which is relatively inaccurate but several orders of magnitude cheaper, and more interpretable under the guidance of 
domain expertise and knowledge priors. (2) point cloud based \cite{YinCenterPoint,zhou2018voxelnet, yan2018second,
xu2020behind,lang2019pointpillars,ciassd, Zheng_2021_CVPR,voxelrcnn,Sheng_2021_ICCV, Mao_2021_ICCV}, which has a 
relatively higher accuracy and lower latency but more prohibitive deployment cost compared with its image based 
counterparts. (3) multimodal fusion based \cite{PointPainting2020Vora, chen2017multi,ku2018avod, ContFuse2018ming,
3dcvf2020jin,CLOCs2020}, which currently lags behind its point cloud based counterparts but importantly provides 
a redundancy to fall-back onto in case of a malfunction or outage.
}
    
\qr{
Fig. \ref{subfig:comparision} presents a summary showing how this survey differs from existing ones on 3D object 
detection. 3D object detection itself is an algorithmic problem, whereas involved with autonomous driving makes it an 
application issue. As of this main text in June 2021, we notice that rather few investigations \cite{rahman2019recent,
guo2019deep, arnold2019survey, feng2020deep} have looked into this application issue. Survey \cite{rahman2019recent} 
focuses on 3D object detection, also taking into account indoor detection. Survey \cite{feng2020deep} involves with 
autonomous driving but it concentrates on multi-modal object detection.} Survey \cite{guo2019deep} covers a series 
of related subtopics of 3D point clouds (\eg, 3D shape classification, 3D object detection and tracking, and 3D 
point cloud segmentation \etc). \qr{Note that survey \cite{guo2019deep} establishes it taxonomy based on network 
architecture, which fails to summarize the homogeneous properties among methods and therefore results in 
overlapping in the subdivisions, \eg multi-view based and BEV based are the same in essence in terms of learning 
strategies. As far as we know, only survey \cite{arnold2019survey} is closely relevant to this paper, but it 
fails to track the latest datasets (\eg, nuScenes \cite{nuscenes2019}, and Waymo Open \cite{Sun_2020_CVPR}), 
algorithms (\eg, the best algorithm it reports on KITII \cite{geiger2012we} 3D detection benchmark is 
AVOD \cite{ku2018avod}[IROS$'$18: 71.88] vs. BtcDet \cite{xu2020behind}[AAAI$'$22: 82.86] in this paper), and 
challenges, which is not surprising as much progress has been made after 2018. 
}                                 

\qr{The aims of this paper are threefold. First, notice that no recent literature exists to collect the growing 
knowledge concerning 3D object detection,} we aim to fill this gap by starting with several basic concepts, 
providing a glimpse of evolution of 3D object detection, together with comprehensive comparisons on publicly 
available datasets being manifested, with pros and cons being judiciously presented. 
\qr{Witnessing the absence of a universal consensus on taxonomy with respect to 3D object detection, our second 
goal is to contribute to the maturity of the taxonomy. To this end, we are cautiously favorable to the 
taxonomy based on input modality, approved by existing literature \cite{rahman2019recent, arnold2019survey}. 
The idea of grouping literature based on their network architecture derives from 2D object detection, which 
fails to summarize the homogeneous properties among methods and therefore results in overlapping in the 
subdivisions, \eg, multi-view based and BEV based are the same representation in essence. Another drawback is 
that several plug-and-play module components can be integrated into either region proposal based (two-stage) 
or single shot based (one-stage). For instance, VoTr \cite{Mao_2021_ICCV} proposes voxel transformer which can be 
easily incorporated into voxel based one stage or two stage detectors. Notice that diverse fusion variants 
consistently emerge among 3D object detection, existing taxonomy in works \cite{rahman2019recent, arnold2019survey} 
needs to be extended. For instance, works \cite{wang2019frustum, wang2019frustum} are sequential fusion methods 
(see Sec. \ref{subsubsec:SequentialFusionbasedMethods}), which are not well suited to existing taxonomy. We therefore 
define two new paradigms, \ie sequential fusion and parallel fusion, to adapt to underlying changes and further 
discuss which category each method belongs to explicitly, while works \cite{rahman2019recent, arnold2019survey} not. 
Also, we analyze in deep to provide a more fine-grained taxonomy above and beyond the existing efforts \cite{rahman2019recent, 
arnold2019survey} on image based methods, say, \emph{result-lifting based} and \emph{feature-lifting based} depending 
on intermediate representation existence.} Finally, we open point-voxel branch to classify newly proposed variants, 
\eg PV-RCNN \cite{pvrcnn2020}. By constrast, survey \cite{guo2019deep} directly groups PV-RCNN into ``Other Methods'', 
leaving the problem unsolved. \qr{Our third goal aims to present a case study on fifteen selected models among surveyed 
works, with regard to runtime analysis, error analysis, and robustness analysis closely. We argue that what mainly 
restricts the performance of detection is 3D location error based on our findings. Taken together, this survey is 
expected to foster more follow-up literature in 3D vision community.
}

The rest of this paper is organized as follows. Section \ref{sec:background} introduces background associated with 
foundations, sensors, datasets, and performance metrics. Section \ref{sec:approach} reviews 3D object detection methods 
with their corresponding pros and cons in the context of autonomous driving. Comprehensive comparisons of the 
state-of-the-arts are summarized in Section \ref{sec:eval}. We conclude this paper and identify future research 
directions Section \ref{sec:con}. \qr{We also set up a regularly updated project page on:
\href{https://github.com/rui-qian/SoTA-3D-Object-Detection}{\emph{https://github.com/rui-qian/SoTA-3D-Object-Detection}}.
}
\vspace{-0.3cm}

\vspace{0.5em}
\section{BACKGROUND} \label{sec:background}
  
\subsection{Foundations} \label{Foundations}
\qr{
Let $\mathcal{X}$ denote input data, say, LiDAR signals or monocular images, $\mathcal{F}$ denote a detector 
parameterized by $\Theta$. Consider an $(F+1)$-dimensional result subset with $n$ predictions, denoted by 
$\left\{ \mathbf{y}_1,...,\mathbf{y}_{n} \right\} \subseteq \mathbb{R}^{F+1}$, we have
\begin{equation}
  \begin{aligned}        
    \Theta _{MLE}=\underset{\Theta}{\text{arg}\max}\mathcal{F}\left( \left. \left\{ \mathbf{y}_1,...,\mathbf{y}_{n} \right\}\ \right|\ \mathcal{X},\ \Theta \right),
    \label{eq:Foundations}
  \end{aligned}      
\end{equation}
where $\mathbf{y}_i=\left( \mathcal{B}_i,s_i \right)$ denotes a certain prediction of detector $\mathcal{F}\left( \mathcal{X}\ ;\Theta \right) $
with bounding box $\mathcal{B}_i\in \mathbb{R}^F$ and its probabilistic score $s_i\in \left[ 0,1 \right]$.
In the context of autonomous driving,  $\mathcal{B}_i$ is usually parameterized as portrayed 
in Fig. \ref{subfig:corners}, which indicates the volume of the object of interest and 
its position relative to a reference coordinate system that can be one of the sensors equipped on a 
ego-vehicle. We notice that attributes encoded by (d) in Fig. \ref{subfig:corners} are orthogonal and 
therefore result in a more lower information redundancy compared with (a), (b), (c). In this paper, 
we adopt the form of (d) as most previous works \cite{yan2018second,zhou2018voxelnet,lang2019pointpillars,
shi2019pointrcnn,Weng_2019_ICCV_Workshops} do.
\begin{figure}[pos=htbp] 
    \setlength{\abovecaptionskip}{0.1cm}
    \setlength{\belowcaptionskip}{0cm}
        \centering  
          \includegraphics[width=3.3in,height=1.3in]{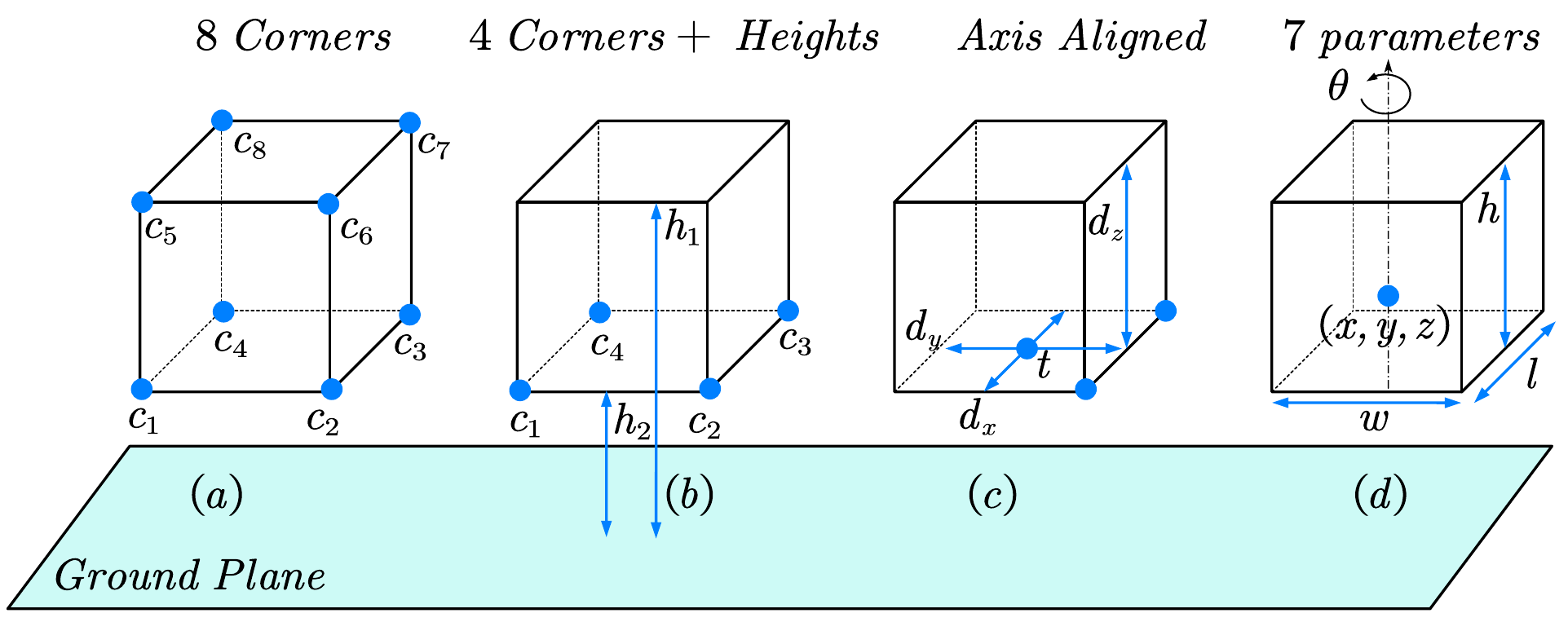} 
          \caption{\textbf{Comparisons of the 3D bounding box parameterization}, between 8 corners proposed 
          in \cite{chen2017multi}, 4 corners with heights proposed in \cite{ku2018avod}, the axis aligned box 
          encoding proposed in \cite{dssa3d2016song}, and the 7 parameters for an oriented 3D bounding box adopted 
          in \cite{yan2018second,zhou2018voxelnet,lang2019pointpillars,shi2019pointrcnn,Weng_2019_ICCV_Workshops}.}
          \label{subfig:corners} 
     \vspace{-0.5cm}   
\end{figure}
}

\subsection{Sensors} \label{Sensors}

We human beings leverage visual and auditory systems to perceive the real world when driving, so how about 
autonomous vehicles? If they were to drive like a human, then to identify what they see on the road constantly 
is the way to go. To this end, sensors matter. It is sensors that empower vehicles a series of abilities:  
obstacles perception, 
automatic emergency braking, and collision avoidance \etc.
In general, the most commonly used sensors can be divided into two categories: passive sensors and active sensors 
\cite{Whatarepassiveandactivesensors}. The on going debate among industry experts is whether or not to just equip 
vehicles with camera systems (no LiDAR), or deploy LiDAR together with on-board camera systems. 
Given that camera is considered to be one of the typical representatives of passive sensors, and LiDAR is regarded 
as a representative of active ones, we first introduce the basic concepts of passive sensors and active sensors, 
then take camera and LiDAR as examples to discuss how they serve the autonomous driving system, together with pros 
and cons being manifested in Table \ref{tab:sensors}.
\vspace{5pt}
\begin{table*}[htbp]
    \setlength{\abovecaptionskip}{0.1cm}
    \setlength{\belowcaptionskip}{0.2cm}
    \newcommand{\tabincell}[2]{\begin{tabular}{@{}#1@{}}#2\end{tabular}} 
    \centering 
    \caption{\textbf{Advantages and disadvantages of different sensors.}}
    \label{tab:sensors}
    \resizebox{\textwidth}{!}{
    \begin{tabular}{m{0.1\linewidth}m{0.15\linewidth}p{0.37\linewidth}p{0.36\linewidth}p{0.15\linewidth}}
    \toprule
     & Sensors     & Advantages          & Disadvantages    & Publications\\
    \midrule
    \multirowcell{5}{Passive}&
    \multirowcell{1}{Monocular Camera}
                                   &\parbox{\linewidth}{  \textbullet cheap and available for multiple situations \\
                                                        \textbullet informative color and texture attributes }     
                                   &  \parbox{\linewidth}{ \textbullet no depth or range detecting feature \\ 
                                                        \textbullet susceptible to weather and light conditions}   
                                   & \parbox{\linewidth}{ \cite{mono3d2016chenxiaozhi}, \cite{ChabotCRTC17DeepMANTA}, \cite{MousavianAFK17Deep3DBox}, \cite{XuC18mf3d}, \cite{li2019gs3d}, \cite{Weng_2019_ICCV_Workshops}}\\
    \cmidrule{2-5}
    &\multirowcell{1}{Stereo Camera}
                                   &  \parbox{\linewidth}{ \textbullet depth information provided \\ \textbullet informative color and texture attributes \\   \textbullet high frame rate}                  
                                   &  \parbox{\linewidth}{ \textbullet computationally expensive  \\ \textbullet sensitive to weather and light conditions  \\ \textbullet limited Field-of-View} 
                                   & \parbox{\linewidth}{ \cite{li2019stereo}, \cite{3dop2015chennips}, \cite{3dop2018chen} } \\
    \midrule
    \multirowcell{6}{Active}&
    \multirowcell{1}{LiDAR}
                                  &  \parbox{\linewidth}{ \textbullet  accurate depth or range detecting feature  \\ \textbullet less  affected  by external illumination \\ \textbullet $360^\circ$ Field-of-View  } 
                                  &  \parbox{\linewidth}{ \textbullet   high sparseness and irregularity by nature\\  \textbullet  no color and texture attributes  \\ \textbullet expensive and critical deployment }
                                  & \parbox{\linewidth}{\cite{VeloFCN2016li}, \cite{yang2018pixor}, \cite{ pvrcnn2020}, \cite{shi2019pointrcnn}, \cite{yang2019std}, \cite{Yang2020ssd}, \cite{Point-GNN}, \cite{he2020sassd}, \cite{yan2018second}, \cite{zhou2020end}, \cite{lang2019pointpillars}, \cite{voxelfpn2020Kuang}, \cite{zhou2018voxelnet}, \cite{wang2019frustum}, \cite{shi2020points2parts}, \cite{hvnet2020ye}}\\
    \cmidrule{2-5}
    &\multirowcell{1}{Solid State \\ LiDAR}
                                 & \parbox{\linewidth}{ \textbullet more reliable compared with surround view sensors   \\ \textbullet cost decrease}          
                                 & \parbox{\linewidth}{ \textbullet error increase when different points of view are merged in real time \\ \textbullet still under development and limited Field-of-View }  
                                     & \tabincell{c}{n.a.}\\
    \bottomrule
    \end{tabular}}
\end{table*}
\begin{table*}[htbp] 
  \setlength{\tabcolsep}{5pt}
  \centering
  \renewcommand\arraystretch{1.2}
  \caption{\textbf{A summary of publicly available datasets for 3D object detection in the context of autonomous driving.}
  *: Numbers in brackets indicate classes evaluated in their official benchmarks.}
  \newcommand{\tabincell}[2]{\begin{tabular}{@{}$\sharp$1@{}}$\sharp$2\end{tabular}} 
 \resizebox{\textwidth}{!}{
    \begin{tabular}{rrrrrrcccccccccccc}
    \toprule
    \multirow{2}*{\textbf{Dataset}} &\multirow{2}*{\textbf{Year}} &\multicolumn{5}{c}{\textbf{Size}} & & \multicolumn{4}{c}{\textbf{Diversity}} & & \multicolumn{3}{c}{\textbf{Modality}} & \multirow{2}*{\textbf{Benchmark}} & \multirow{2}*{\textbf{Cites}}\\
   \cline{3-7} \cline{9-12} \cline{14-16}     
     &  & \multirowcell{1}[0pt][c]{\textbf{$\sharp$Train}} &\multirowcell{1}[0pt][c]{\textbf{$\sharp$Val}} & \multirowcell{1}[0pt][c]{\textbf{$\sharp$Test}} & \multirowcell{1}[0pt][c]{\textbf{$\sharp$Boxes}} & \multirowcell{1}[0pt][c]{\textbf{$\sharp$Frames}} & & \multirowcell{1}[0pt][c]{\textbf{$\sharp$Scenes}} & \multirowcell{1}[0pt][c]{\textbf{$\sharp$Classes*}} & \multirowcell{1}[0pt][c]{\textbf{Night}} & \multirowcell{1}[0pt][c]{\textbf{Rain}} & & \multirowcell{1}[0pt][c]{\textbf{Stereo}} & \multirowcell{1}[0pt][c]{\textbf{Temporal}} & \multirowcell{1}[0pt][c]{\textbf{LiDAR}} &  \\
     \midrule
     KITTI \cite{geiger2012we,geiger2013vision}     & 2012  & 7,418$\times$1 & - & 7,518 $\times$ 1  & 200K & 15K & & 50 & 8 (3) & No & No & & Yes & Yes & Yes & Yes & 5011 \\
     Argoverse \cite{nuscenes2019}    & 2019  & 39,384$\times$7 & 15,062$\times$7 & 12,507 $\times$ 7  & 993K  & 44K & & 113 & 15 & Yes & Yes & & Yes & Yes & Yes & Yes & 88\\
     Lyft L5 \cite{lyft2019av}    & 2019  & 22,690$\times$6 & - & 27,460 $\times$6   & 1.3M & 46K & & 366 & 9 & No & No & & No & Yes & Yes & No & - \\
     H3D \cite{360LiDARTracking_ICRA_2019}    & 2019  & 8,873$\times$3 & 5,170$\times$3 & 13,678 $\times$3   & 1.1M & 27K & & 160 & 8 & No & No & & No & Yes & Yes & No & 31\\
     Appllo \cite{ma2019trafficpredict,Huangtpami2020}    & 2019  & - & - & -   & - & 140K & & 103 & 27 & Yes & Yes & & Yes & Yes & Yes & Yes & 78\\
     nuScenes \cite{nuscenes2019}    & 2019  & 28,130$\times$6 & 6,019$\times$6 & 6,008 $\times$6   & 1.4M & 40K & & 1,000 & 23 (10) & Yes & Yes & & No & Yes & Yes & Yes & 225\\
     Waymo \cite{Sun_2020_CVPR}    & 2020  & 122,200$\times$5 & 30,407$\times$5 & 40,077 $\times$5   & 112M & 200K & & 1,150 & 4 (3) & Yes & Yes & & No & Yes & Yes & Yes &31\\
     \bottomrule									
  \end{tabular}%
  \label{tab:datasets}
  }
\end{table*}%
\subsubsection{Passive Sensors}\label{sec:PassiveSensors}
Passive sensors are anticipated to receive natural emissions emanating from both the Earth's surface and its atmosphere.
These natural emissions could be natural light or infrared rays. Typically, a camera directly grabs a bunch of 
color points from the optics in the lens and arranges them into an image array that is often referred to as a digital 
signal for scene understanding. 
Primarily, a monocular camera lends itself well to informative color and texture attributes, better visual recognition 
of text from road signs, and high frame rate at a negligible cost \etc. Whereas, it is lack of depth information,  
which is crucial for accurate location estimation in the real 3D world. To overcome this issue, stereo cameras use 
matching algorithms to align correspondences in both left and right images for depth recovery \cite{psmnet2018}. 
While cameras have shown potentials as a reliable vision system, it is hardly sufficient as a standalone one 
given that a camera is prone to degrade its accuracy in cases where luminosity is low at night-time or rainy 
weather conditions occur. \qr{Consequently equipping autonomous vehicles with an auxiliary sensor, say active 
counterparts, to fall-back onto is necessary, in case that camera system should malfunction or disconnect in 
hazardous weather conditions.}

\subsubsection{Active Sensors} \label{ActiveSensors}
Active sensors are expected to measure reflected signals that are transmitted by the sensor, which are bounced by 
the Earth's surface or its atmosphere. Typically, Light Detection And Ranging (LiDAR) is a point-and-shoot device 
with three basic components of lens, lasers and detectors, which spits out light pulses that will bounce off the 
surroundings in the form of 3D points, referred to as ``point clouds''.  High sparseness and irregularity 
by nature and the absence of texture attributes are the primary characteristics of a point cloud, which is well 
distinguished from image array. Since we have already known how fast light travels, the distance of obstacles 
could be determined without effort. LiDAR system emits thousands of pulses that spin around in a circle per second, 
with a 360 degree view of surroundings for the vehicles being provided. For example, Velodyne HDL-64L 
produces 120 thousand points per frame with a 10 Hz frame rate. Obviously, LiDAR is less affected by external 
illumination conditions (\eg, at night-time), given that it emits light pulses by itself. 
Although LiDAR system has been hailed for high accuracy and reliability compared with camera system, 
it does not always hold true. Specifically, wavelength stability of LiDAR is susceptible to variations 
in temperature, while adverse weather (\eg, snow or fog) is prone to result in poor signal-to-noise ratio in 
the LiDAR's detector. Another issue with LiDAR is the high cost of deployment. 
A conservative estimate according to Velodyne, so far, is about \$75,000\footnote[1]{\href{http://www.woodsidecap.com/wp-content/uploads/2018/04/Yole_WCP-LiDAR-Report_April-2018-FINAL.pdf}{http://www.woodsidecap.com}} \cite{arnold2019survey}. 
In the foreseeable future of LiDAR, how to decrease cost and how to increase resolution and range are where the whole 
community is to march ahead. As for the former, the advent of solid state LiDAR is expected to address this problem 
of cost decrease, with the help of several stationary lasers that emit light pulses along a fixed field of view. 
As for the latter, the newly announced Velodyne VLS-128 featuring 128 laser pulses and 300m radius range has been 
on sale, which is going to significantly facilitate better recognition and tracking in terms of public safety.  

\begin{table*}[htbp]
  \centering
  \caption{\textbf{Instance distribution on nuScenes \emph{train} split.}
  Here, ``TC'' and ``Cons.Veh.'' denote traffic cone and construction vehicle respectively.}
  \setlength{\tabcolsep}{0.9mm}{
    \begin{tabular}{ccccccccccc}
    \toprule
    $\sharp$\textbf{Classes} & \textbf{Car}   & \textbf{Pedestrian} & \textbf{Barrier} & \textbf{TC} & \textbf{Truck} & \textbf{Trailer} & \textbf{Bus}   & \textbf{Cons.Veh.} & \textbf{Motocycle} & \textbf{Bicycle} \\
    \midrule
    Number & 413,318 & 185,847 & 125,095 & 82,362 & 72,815 & 20,701 & 13,163 & 11,993 & 10,109 & 9,478 \\
    \bottomrule
    \end{tabular}%
  \label{tab:imbalancenuscenes}}%
  \vspace{-1em}
\end{table*}%

\begin{table*}[!htbp] 
  \centering
  \caption{\textbf{Instance distribution on KITII \emph{train} split.}
  \emph{Car} category accounts for 82.99\% of the three (\ie, \emph{Car}, \emph{Pedestrian}, and \emph{Cyclist}).}
  \setlength{\tabcolsep}{0.9mm}{
    \begin{tabular}{ccccccccc}
    \toprule
    $\sharp$\textbf{Classes} & \textbf{Car} & \textbf{Ped.}   & \textbf{Van}   & \textbf{Cyclist} & \textbf{Truck} & \textbf{Misc}  & \textbf{Tram}  & \textbf{Person.sit.} \\
    \midrule
    Number & 14,357  & 2,207 & 1,297  & 734   & 448   & 337   & 224   & 56 \\
    \bottomrule
    \end{tabular}%
  \label{tab:imbalancekitti}
  \vspace{-1em}
  }
\end{table*}%
\subsubsection{Discussion} 
Fatalities occurred with autonomous vehicles have already increased the society's grave concern about safety. 
If autonomous vehicles were to hit the road legally, they at least need to satisfy three basic requirements: 
high accuracy, high certainty, and high reliability. To this end, sensor fusion incorporating the merits of 
two worlds (camera vs. LiDAR) is going to be necessary. From a sensor standpoint, LiDAR provide depth information 
close to linearity error with a high level of accuracy, but it is susceptible to adverse weather (\eg, snow or fog). 
Camera is intuitively much better at visual recognition in cases where color or texture attributes are available, 
but they are not sufficient as a standalone system as aforementioned. 
Note that certainty is still an important yet largely unexplored problem. A combination of LiDAR and camera is 
anticipated to ensure detection accuracy and improve prediction certainty. With regard to reliability, two facets 
should be considered: sensor calibration and system redundancy. Sensor calibration undoubtedly increases the 
difficulty of deployment and directly affects the reliability of the whole system. Studies \cite{HALLORAN2020107058, DORNAIKA20072716} 
have looked into calibrating sensors to avoid drift over time. System redundancy is to have a secondary sensor 
to fall-back onto in case of a malfunction or outage in extreme scenarios. \qr{Although balancing affordability and 
safety has been a long-term ethical dilemma}, the community should be keenly aware of the safety risk of over-reliance 
on a single sensor. 

\subsection{Dataset}
The availability of large-scale datasets has been continuously fostering the community 
with the advent of data-driven era. As regards 3D object detection, we summarize publicly 
available datasets \cite{geiger2012we,geiger2013vision, nuscenes2019, lyft2019av, 360LiDARTracking_ICRA_2019, 
ma2019trafficpredict, Huangtpami2020, Argoverse, Sun_2020_CVPR} in the context of autonomous driving in 
Table \ref{tab:datasets}, \qr{out of which the KITTI \cite{geiger2012we}, nuScenes \cite{nuscenes2019}, 
and Waymo Open \cite{Sun_2020_CVPR} are the typical representatives. In a sequel, we selectively 
introduce these three datasets with regard to their size, diversity, pros and cons accordingly.
}

\textbf{Dataset size.} The KITTI manually annotates 200K boxes among 15K frames, with 7,481, 7,518 
samples for training and testing respectively. Rather, the training set is subdivided into 3,712 and 
3,769 samples for \emph{train}, \emph{val} split as a common practice initially introduced by 3DOP \cite{chen2015nips3dop}. 
\qr{The nuScenes manually labels 1.4M boxes among 40K frames, with 28,130, 6,019 and 6,008 frames for 
training, validation, and testing accordingly. Waymo Open, encouragingly annotates 112M boxes among 
200K frames, with 122,200, 30,407 and 40,077 for training, validation, and testing. Note that only 
the labels for training/validation are available, whereas none of them provide annotations for testing. 
Competitors are required to submit predictions to the online leaderboard for fairly assessing on 
\emph{test} set.
}

\textbf{Diversity.} The KITTI provides 50 scenes over 8 classes in Karlsruhe, Germany, out of which 
only \emph{Car}, \emph{Pedestrian}, and \emph{Cyclist} are considered for online evaluation. Three 
difficulty levels (\ie, \emph{Easy}, \emph{Moderate}, and \emph{Hard}) for its protocol are introduced 
depending on the height of 2D bounding boxes, the level of occlusion and truncation. \qr{The nuScenes 
collects 1,000 sequences over 23 classes in Boston and Singapore, out of which only 10 classes are 
considered for 3D object detection. The Waymo Open consists of 1,150 sequences over 4 classes 
in Phoenix and San Francisco, out of which only three classes are assessed similar to KITTI. It is 
worth mentioning that both nuScenes and Waymo Open acquire their data under multiple weather 
(\eg, rainy, foggy, and snowy \etc) and lighting (\eg, daytime and nighttime) conditions throughout 
a day, whereas KITTI only captures its dataset on sunny days.
}

\qr{
\textbf{Pros and cons.} 
These three datasets manifested above indeed catalytically foster the academics. KITTI, as a pioneer, 
has profoundly influences the whole community in terms of data acquisition, protocol and benchmark. 
Nevertheless, as we mentioned above, KITTI is recorded in daytime on sunny days, without taking 
lighting and weather conditions into account, resulting a relative lower diversity compared with 
nuScenes and Waymo Open. Real dataset tends to suffer from class imbalance as it is true to life. 
According to our statistics in Table \ref{tab:imbalancenuscenes}, 50\% categories account for only 6.9\% 
of total annotations, which clearly reflects a long-tail distribution on nuScenes. This phenomenon also holds true 
for KITTI as indicated in Table \ref{tab:imbalancekitti}.
}
\begin{figure*}[pos=htbp]
  \setlength{\abovecaptionskip}{0.1cm}
  \setlength{\belowcaptionskip}{0cm}
    \centering     
        \includegraphics[width=0.95\textwidth]{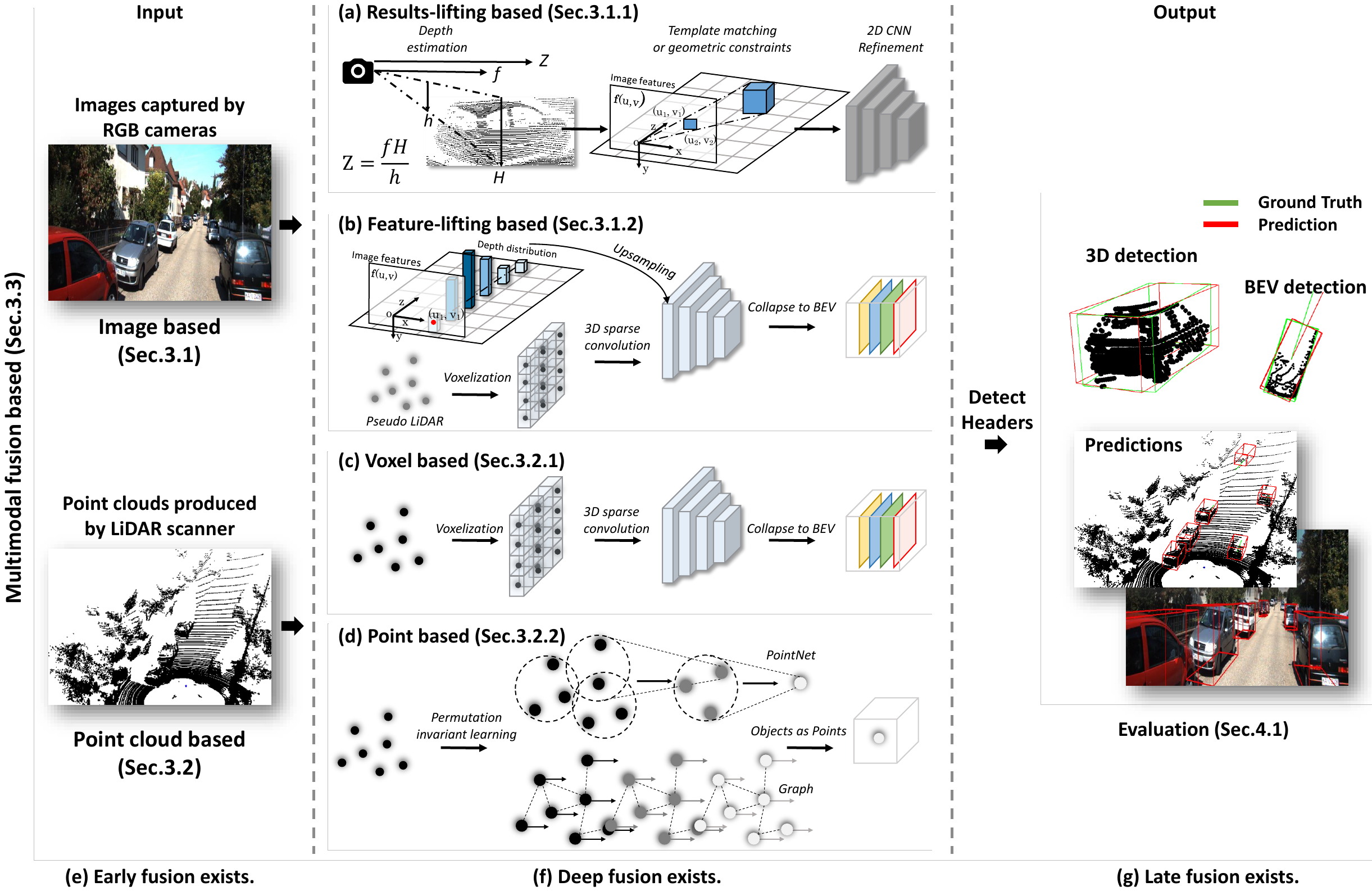} 
        \caption{\textbf{Pipeline of 3D object detection in general.} 
        \emph{Image based}, which either lifts estimated 2D results into 3D space via template matching, geometric constraints 
        in (a), or directly lifts 2D image features into 3D space via computing a Pseudo LiDAR, learning a latent depth 
        distribution in (b). 
        \emph{Point clouds based}, which either voxelizes an irregular point cloud into regular voxel grids 
        and then learn feature representation in an explicit way in (c), or leverage PointNet-like block, GNNs to learn 
        permutation-invariant representations in an implicit fashion in (d). 
        \emph{Multimodal fusion based}, which is likely 
        to fuse cross-modalities at early phase in (e), middle phase in (f), and late phase in (g) during the forward 
        propagation.
        }
    \label{subfig:surveyoverview} 
   \vspace{-1em}
\end{figure*}
\subsection{Performance Metrics}
3D object detection adopts the Average Precision ($AP$) as its primary performance metrics, all of which stem from 
the same ideology as its 2D counterparts \cite{pascal-voc-2012}. We first revisit the vanilla form of 
$AP$ metric, and then recognize subtle connections and differences of dataset-specific $AP$ adopted among commonly 
used benchmarks.

\qr{
\textbf{Revisiting.}
Consider a prediction subset $\left\{ \mathbf{y}_1,...,\mathbf{y}_{n} \right\}$ in descending order by 
confidence score $s_i$, a prediction $\mathbf{y}_i$ is considered as true positive if the ratio of overlapping 
area between $\mathcal{B}_i$ and its assigned ground-truth, namely, the Intersection over Union (IoU), 
exceeds a certain threshold, otherwise false positive. The zigzag-like precision-recall curve can easily 
be plotted and $AP$ is just the area under the curve. Notice that accurately calculating the related area 
is difficult, PASCAL VOC \cite{pascal-voc-2012} established an alternative instead. 
}

\emph{Interpolated $AP|_{R_N}$ Metric} is formulated as the mean precision calculated at
a recall subset $R$ of which $N$ evenly spaced recall levels are composed, that is
\begin{equation}
  \begin{aligned}        
      AP|_{R_N}=\frac{1}{N}\sum_{r\in R}{P_{interpolate}\left( r \right)},
      \label{eq:AP}
  \end{aligned}      
\end{equation} 
where $R=\left[ r_0,r_0+\frac{r_1-r_0}{N-1},r_0+\frac{2\left( r_1-r_0 \right)}{N-1},...,r_1 \right]$. 
For each recall level $r$, its corresponding precision is \emph{interpolated} by taking into account the 
maximum precision of which recall value is greater than or equal to $r$, denoted by 
\begin{equation}
  \begin{aligned}        
      P_{interpolate}\left( r \right) =\underset{\tilde{r}:\tilde{r}\ge r}{\max}\ P\left( \tilde{r} \right).
      \label{eq:1}
  \end{aligned}      
\end{equation} 

\textbf{KITTI Benchmark.} KITTI adopts standard \emph{Interpolated $AP|_{R_{11}}$ Metric} as its official metric 
for assessing detector $\mathcal{F}\left( \mathcal{X}\ ;\Theta \right) $. Usually two leaderboard is considered, 
\ie 3D detection and Bird's Eye View (BEV) detection. 3D detection evaluates ${AP}_{3D}|_{R_{11}}$ with a rotated $IoU_{3D}$ 
threshold of 0.7, 0.5, 0.5 for \emph{Car}, \emph{Pedestrian} and \emph{Cyclist} accordingly. The principles of 3D 
detection largely holds true for BEV detection, except for the calculation of ${IoU}_{BEV}$, which is calculated  
by projecting the bounding box $\mathcal{B}_{3D}$ from 3D pace into the ground plane. Note that from 08.10.2019,  
KITTI turns to 40 recall levels $\left[ 1/40,2/40,3/40,...,1 \right]$ instead of 11 recall levels 
$\left[ 0,1/10,2/10,...,1 \right]$ as suggested in \cite{SimonelliBPLK19}, with recall level 0 being reasonably 
removed.

\qr{
\textbf{nuScenes Benchmark.} nuScenes uses NuScenes Detection Score (NDS) as 
its official metric for evaluating detector $\mathcal{F}\left( \mathcal{X}\ ;\Theta \right) $. Consider a mean 
average error subset $\mathcal{E}$ of which translation, size, orientation, attribute and velocity are composed, 
denoted by $\mathcal{E}=\{ mATE,\ mASE,\ mAOE,\ mAAE$ $,\ mAVE \}$, we have 
\begin{equation}
	\begin{aligned}        
		NDS=\frac{1}{10}\left[ 5mAP+\sum_{err\in \mathcal{E}}{\left( 1-\min \left( 1,\ err \right) \right)} \right],
		\label{eq:1}
	\end{aligned}      
\end{equation}        
where $mAP$ indicates mean Average Precision. NDS jointly justifies a weighted average of $mAP$ and mean average errors 
of set $\mathcal{E}$ among 10 classes. It is worth noting that $mAP$ is calculated by a bird-eye-view 
center distance of thresholds $\{0.5m, 1m, 2m, 4m\}$ rather than standard box-overlap.
}

\qr{
\textbf{Waymo Benchmark.} Waymo Open leverages \emph{Interpolated} $AP|_{R_{21}}$ \emph{Metric} and Average
Precision weighted by Heading (APH) as its primary metric for evaluating detector $\mathcal{F}\left( \mathcal{X}\ ;\Theta \right) $. 
To compute $AP$, Waymo evaluates on 21 equally spaced recall levels $\left[ 0,1/20,2/20,...,1 \right]$ in 
Equation \ref{eq:AP} with an IoU threshold of 0.7, 0.5 for vehicles, pedestrians respectively. To compute APH, 
the heading accuracy is incorporated into true positives, each of which is weighted by 
$\min \left( \left| \theta -\theta ^* \right|,\ 2\pi -\left| \theta -\theta ^* \right| \right) /\pi$, where  
$\theta$ and $\theta ^*$ are subject to $\left[ -\pi ,\pi \right] $, indicating the predicted azimuth and 
its assigned ground truth accordingly. Waymo breaks down its difficulty into two levels: $\text{LEVEL\_}1$ lends itself to 
boxes with at least five LiDAR signals, while $\text{LEVEL\_}2$ is suited for all non-empty ones. 
}

\vspace{0.5em}
\section{TAXONOMY AND REVIEW} \label{sec:approach}
\begin{table*}[htbp] 
  \centering
  \renewcommand\arraystretch{1.25}
  \caption{\textbf{A taxonomy of 3D object detection for autonomous driving.}}
  \newcommand{\tabincell}[2]{\begin{tabular}{@{}#1@{}}#2\end{tabular}} 
 \resizebox{\textwidth}{!}{
    \begin{tabular}{m{0.1\textwidth}m{0.15\textwidth}<{\centering}p{0.35\textwidth}p{0.4\textwidth}}
    \toprule
    \multirow{2}*{ \tabincell{c}{\textbf{Input} \\\textbf{modality}}} &\multirow{2}*{\textbf{Subdivision}} & \multicolumn{2}{c}{\textbf{Network architecture}}\\
 \cline{3-4}    
     &  & \multirowcell{1}[0pt][c]{\textbf{Two-stage}} &\multirowcell{1}[0pt][c]{\textbf{One-stage}}  \\
    \midrule
     \multirow{14}*{ \tabincell{c}{\textbf{Image} \\(Sec.\ref{subsec:Image-based})} } & \multicolumn{3}{l}{\it{Depending on intermediate representation existence:}}\\
     \cline{2-4}  
                  & \multirow{8}*{\makecell{result-lifting \\(Sec.\ref{subsubsec:ResultliftingbasedMethods})}} &   {\it with extra data:} &{\it with extra data:} \\
                  & & + Mono3D \cite{mono3d2016chenxiaozhi}, CVPR$'$16 & + DD3D \cite{Park_2021_ICCV}, ICCV$'$21\\
                  & &  {\it without extra data:} &{\it without extra data:} \\
                  & & + Deep MANTA \cite{ChabotCRTC17DeepMANTA}, CVPR$'$17 & + Monodle \cite{Ma_2021_CVPR}, CVPR$'$21\\
                  & & + Deep3DBox \cite{MousavianAFK17Deep3DBox}, CVPR$'$17 & + MonoFlex \cite{Zhang_2021_CVPR}, CVPR$'$21\\
                  & & + GS3D \cite{li2019gs3d}, CVPR$'$19 & + YOLOStereo3D \cite{2021YOLOStereo3D}, ICRA$'$21\\
                  & & + Stereo R-CNN \cite{li2019stereo}, CVPR$'$19 & \\
                  & & + MonoRCNN \cite{Shi_2021_ICCV}, ICCV$'$21 & \\                
\cline{2-4}
                  & \multirow{8}*{\makecell{feature-lifting \\(Sec.\ref{subsubsec:FeatureliftingbasedMethods})}} &   {\it with extra data:} &{\it with extra data:} \\
                  & &  + MF3D \cite{XuC18mf3d}, CVPR$'$18 & + OFT-Net \cite{OFT-NetBMVC}, BMVC$'$19\\
                  & & + Mono3D-PLiDAR \cite{Weng_2019_ICCV_Workshops},ICCVW$'$19 & + DSGN \cite{Chen0SJ20}, CVPR$'$20\\
                  & & + Pseudo-LiDAR \cite{wang2019pseudo}, CVPR$'$19& + LIGA-Stereo \cite{Guo_2021_ICCV}, ICCV$'$21\\
                  & & + Pseudo-LiDAR++ \cite{wang2019pseudo++}, ICLR$'$20 & + CaDDN \cite{Reading_2021_CVPR}, CVPR$'$21\\
                  & & + Pseudo-LiDAR E2E \cite{qian2020e2e},CVPR$'$20 & \\
                   & &  {\it without extra data:} & \\     
                  & & + 3DOP \cite{3dop2018chen}, NeurlPS$'$15 & \\								
\hline										
     \multirow{20}*{ \tabincell{c}{\textbf{Point cloud} \\(Sec.\ref{subsec:Point-Cloud-based})} } & \multicolumn{3}{l}{\it{Depending on representation learning strategies:}}\\
     \cline{2-4} 
                  & \multirow{14}*{\makecell{voxel based \\(Sec.\ref{subsubsec:VoxelbasedMethods})}} &   {\it single-scale voxelization:} &{\it single-scale voxelization:} \\
                  & & + TANet \cite{TANet2020}, AAAI$'$20 & + VeloFCN \cite{VeloFCN2016li}, RSS$'$16\\
                  & & + SPG \cite{Xu_2021_ICCV}, ICCV$'$21 & + PIXOR \cite{yang2018pixor}, CVPR$'$18\\
                  & & + CenterPoint \cite{YinCenterPoint}, CVPR$'$21 & + VoxelNet \cite{zhou2018voxelnet}, CVPR$'$18\\
                  & & + BtcDet \cite{xu2020behind}, AAAI$'$22 &  + SECOND \cite{yan2018second}, Sensors$'$18 \\           
                  & & & + PointPillars \cite{lang2019pointpillars}, CVPR$'$19\\
                  & & & + CIA-SSD \cite{ciassd}, AAAI$'$21\\   
                  & & & + SE-SSD \cite{Zheng_2021_CVPR}, CVPR$'$21\\
                  & & & + Voxel R-CNN \cite{voxelrcnn}, AAAI$'$21\\
                  & & & + CT3D \cite{Sheng_2021_ICCV}, ICCV$'$21\\
                  & & & + VoTr \cite{Mao_2021_ICCV}, ICCV$'$21\\
                & &    {\it multi-scale voxelization:} &\ {\it multi-scale voxelization:}\\
                   & & + Part-A$^2$ \cite{shi2020points2parts}, T-PAMI$'$21 & + HVNet \cite{hvnet2020ye}, CVPR$'$20\\
                  & & & + Voxel-FPN \cite{voxelfpn2020Kuang}, Sensors$'$20\\
\cline{2-4}
                  &\tabincell{c}{point based \\(Sec.\ref{subsubsec:PointbasedMethods})} &   \tabincell{l}{+ PointRCNN \cite{shi2019pointrcnn}, CVPR$'$19}
                &\tabincell{l}{+ 3DSSD \cite{Yang2020ssd}, CVPR$'$20\\
                  + Point-GNN \cite{Point-GNN}, CVPR$'$20\textcolor{white}{a}}\\
\cline{2-4}
                  &\tabincell{c}{point-voxel based \\(Sec.\ref{subsubsec:PointvoxelbasedMethods})}&   \tabincell{l}{+ Fast PointRCNN \cite{Chen0SJ19}, ICCV$'$19\\
                  + STD \cite{yang2019std}, ICCV$'$19\\
                  + PV-RCNN \cite{pvrcnn2020}, CVPR$'$20\\
                  + BADet \cite{qian_2022_PR}, PR$'$22}
              &     \tabincell{l}{+ SA-SSD \cite{he2020sassd}, CVPR$'$20\textcolor{white}{aaa}}\\		
\hline									
     \multirow{12}*{ \tabincell{c}{\textbf{Multimodal} \\(Sec.\ref{subsec:Multimodal-Fusion-based})} } & \multicolumn{3}{l}{\it{Depending on to what extent these two modalities are coupled:}}\\
     \cline{2-4}
                  &\tabincell{c}{sequential fusion \\(Sec.\ref{subsubsec:SequentialFusionbasedMethods})} &    \tabincell{l}{+ Frustum-PointNets \cite{qi2018frustum}, CVPR$'$18\\
                  + Frustum-ConvNet \cite{wang2019frustum}, IROS$'$19 }
                &     \\
\cline{2-4}
                  &\multirow{9}*{\makecell{parallel fusion \\(Sec.\ref{subsubsec:ParallelFusionbasedmethods})} }&   {\it early fusion:} & \\
                  & & + PointPainting \cite{PointPainting2020Vora}, CVPR$'$20 & \\
                  & & {\it deep fusion:} & {\it deep fusion:}\\
                  & & + MV3D \cite{chen2017multi}, CVPR$'$17 & + ContFuse \cite{ContFuse2018ming}, ECCV$'$18 \\
                  & & + AVOD \cite{ku2018avod}, IROS$'$18 & \\
                  & & + MMF \cite{mmf2019cvpr}, CVPR$'$19 & \\
                  & & + 3D-CVF \cite{3dcvf2020jin}, ECCV$'$20 & \\
                  & & {\it late fusion:} & \\
                  & & + CLOCs \cite{CLOCs2020}, IROS$'$20 & \\                		
  \bottomrule									
  \end{tabular}%
  \label{tab:taxonomy}
  }
\end{table*}%

\qr{    
Fig. \ref{subfig:surveyoverview} presents a pipeline of 3D object detection in general. Having only images on hand, the core challenge arises from the absence of depth information. Usually two lines exist: one is to break 
down this 3D vision task into 2D object detection \cite{girshickICCV15fastrcnn,HeZR014sppnet,RenHGS15fasterrcnn,RedmonDGF16yolo, LiuAESRFB16ssd,LinGGHD17RetinaNet,maskrcnn2017he} 
and depth estimation \cite{psmnet2018, dorn2019fu}, which lifts these estimated results into 3D space 
via geometric properties and constraints. The other line is to directly lift 2D image features into 3D space 
via computing a point cloud \cite{3dop2018chen, XuC18mf3d, Weng_2019_ICCV_Workshops, wang2019pseudo, wang2019pseudo++, 
qian2020e2e}, termed as \emph{Pseudo LiDAR} or learning a latent depth distribution \cite{Chen0SJ20, Guo_2021_ICCV, Reading_2021_CVPR, OFT-NetBMVC}. 
Having only point clouds on hand, the key difficulty stems from the representation learning over sparse, irregular, 
and unordered point clouds. Also two ways exist mainly: one is to first voxelize an irregular point cloud 
into regular voxel grids and then learn feature representations in an explicit way. Nevertheless, the other is to 
leverage PointNet-like block \cite{qi2017pointnet} or Graph Neural Networks (GNNs) \cite{semigcn2017} to learn permutation-invariant 
representations in an implicit fashion. Nowadays, combining the merits of these two lines reveals a new fashion trend. 
What if having both of them on hand? The tricky obstacle often derives from semantic representation (what to fuse), 
alignment (how to fuse), and consistency (when to fuse) for multimodal fusion based on what we have already learnt 
from these two preceding modalities.
}

\qr{Depending on what we feed to  $\mathcal{F}\left( \mathcal{X}\ ;\Theta \right) $ internally during inference}, we 
frame our taxonomy along three dimensions: (1) image based, (2) point cloud based, and (3) multimodal fusion based, 
ordered chronologically in which each method emerges. Table \ref{tab:taxonomy} selectively manifests several core 
literature structured along each dimension, which is expected to leave the readers a clear picture 
in his or her mind. In what follows, we present image based in Section \ref{subsec:Image-based}, 
and point cloud based in Section \ref{subsec:Point-Cloud-based}. Multimodal fusion based is addressed 
in Section \ref{subsec:Multimodal-Fusion-based}.
\vspace{0.53em}
\subsection{Image based Methods} \label{subsec:Image-based}
\vspace{0.1em}
Depth estimation from images is still an ill-posed problem that are not fully understood yet 
\cite{psmnet2018, dorn2019fu}. Errors from depth recovery inherently contribute to the major factor of 
the performance gap between image based and point cloud based \cite{Shi_2021_ICCV}. \qr{Focusing on the inverse 
issue, it is the way of recovering depth and the way of use that collected knowledge that determines how 
the intermediate representation is lifted. Depending on intermediate representation existence, 
we divide this group into the following two subdivisions: (1) result-lifting based, (2) feature-lifting based. 
Table \ref{tab:taxonomy} selectively lists several significant contributions concerning the subject.
}
\vspace{0.43em}
\subsubsection{Result-lifting based Methods} \label{subsubsec:ResultliftingbasedMethods}
\vspace{0.1em}
\qr{
Works in this group break down $\mathcal{F}\left( \mathcal{X}\ ;\Theta \right) $ into two vision tasks: 
2D object detection and depth estimation \cite{ChabotCRTC17DeepMANTA, MousavianAFK17Deep3DBox, li2019gs3d, li2019stereo, 
Shi_2021_ICCV, Park_2021_ICCV, Ma_2021_CVPR, 2021YOLOStereo3D}. The underlying principle is that the spatial location of the 
associated objects can be empirically inferred with regard to the visual appearance.} To that end, 
Mono3D \cite{mono3d2016chenxiaozhi} scores proposals with location prior, object shape, size, and semantics 
with the hypothesis that objects are close to the ground plane via minimizing energy function. Deep3DBox \cite{MousavianAFK17Deep3DBox} 
leverages the geometric properties that the perspective projection of 3D corners should tightly touch at
least one side of the 2D bounding box. GS3D \cite{li2019gs3d} relies on the observation that 
the top center of a 3D bounding box should be close to the top midpoint of the 2D bounding box 
when projected onto the image plane. Stereo R-CNN \cite{li2019stereo} exploits geometric 
alignment with keypoints, yaw angle, and object shape using left and right proposal pairs.
\qr{Literature \cite{Zhang_2021_CVPR, Ma_2021_CVPR, Shi_2021_ICCV} fully exploits semantic properties 
as well as dense geometric constraints in monocular imagery. We notice that these methods are over-reliance on 
feature engineering and hinder them from further extending to general scenarios. For instance, works \cite{mono3d2016chenxiaozhi, ChabotCRTC17DeepMANTA, Park_2021_ICCV} 
require external data for training. Keypoint constraints in works \cite{li2019stereo, Shi_2021_ICCV,Zhang_2021_CVPR} are more likely to be 
vulnerable to slim and tall objects, say, Pedestrians. To what extent works in this group rely on domain 
expertise determines to what extent efforts in this group can be generalized largely. 
}
\vspace{0.43em}
\subsubsection{Feature-lifting based Methods} \label{subsubsec:FeatureliftingbasedMethods}
\vspace{0.1em}
\qr{
Works in this group develop $\mathcal{F}\left( \mathcal{X}\ ;\Theta \right) $ via lifting 2D image features 
into 3D space via computing a point cloud intermediately \cite{3dop2018chen, XuC18mf3d, Weng_2019_ICCV_Workshops, wang2019pseudo, wang2019pseudo++, 
qian2020e2e} or learning a categorical depth distribution directly \cite{Chen0SJ20, Reading_2021_CVPR}.} To this end, 
works \cite{wang2019pseudo, XuC18mf3d, Weng_2019_ICCV_Workshops} first uses a stand-alone depth estimation 
network to obtain disparity map, then back-projects 2D coordinates associated with each pixel in image plane into 
3D space, and finally lends itself to independent point cloud based detectors. Notice that the 
depth estimation error grows quadratically at long ranges, Pseudo-LiDAR++ \cite{wang2019pseudo++} uses 
sparse but accurate real LiDAR signals as ``landmarks'' to correct and de-bias depth errors. 
Observe that preceding networks are trained separately, Pseudo-LiDAR E2E \cite{qian2020e2e} 
goes one step further by making the entire pipeline trainable end to end. We note that Pseudo-LiDAR signals in works 
\cite{wang2019pseudo, wang2019pseudo++, Weng_2019_ICCV_Workshops} are internally accompanied by noises that 
stem from errors in depth estimation, which reflect in two aspects: (1) Pseudo-LiDAR signals are slightly off 
relative to the real LiDAR ones with a local misalignment. (2) depth artifacts commonly exist in the vicinity 
of an object along with a long tail \cite{Weng_2019_ICCV_Workshops}. 
\qr{Works \cite{Chen0SJ20, Reading_2021_CVPR} 
also point out that the independent networks, say depth estimation, make this 3D representation sub-optimal 
concerning the non-differentiable transformation from 2D coordinates into 3D space. Consequently, instead of 
computing an intermediate point cloud, the other scheme wants to incorporate 3D geometry into image based networks 
and learn a latent depth distribution \cite{Chen0SJ20, Guo_2021_ICCV, Reading_2021_CVPR} explicitly or implicitly \cite{OFT-NetBMVC} in an end-to-end manner.
DSGN \cite{Chen0SJ20} and CaDNN \cite{Reading_2021_CVPR} encode visual appearance into 3D feature volumes, jointly 
optimizing depth and semantic cues for 3D detection. LIGA-Stereo \cite{Guo_2021_ICCV} 
turns to real LiDAR signals for the high-level geometry-aware supervisions and beyond under the guidance of 
a teacher network.
}
\begin{figure*}[pos=htbp]
  \setlength{\abovecaptionskip}{0.1cm}
  \setlength{\belowcaptionskip}{0cm}
    \centering     
        \includegraphics[width=0.85\textwidth]{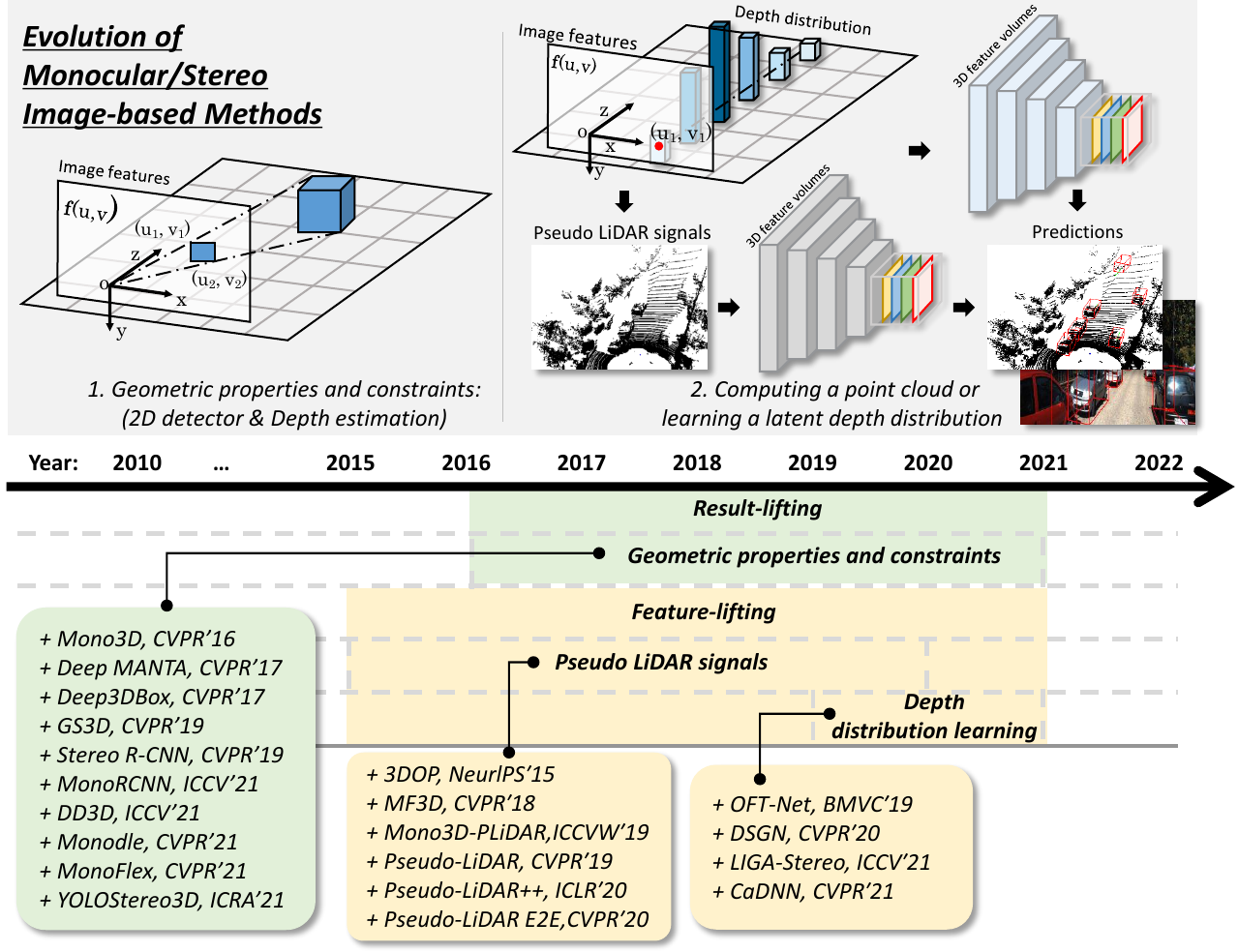} 
        \caption{\textbf{Evolution of monocular or stereo image-based methods.}}
    \label{subfig:imagebasedmethods} 
\end{figure*}
\subsubsection{Summary}
Fig. \ref{subfig:imagebasedmethods} illustrates the evolution of image based. Works in this group are divided 
into result-lifting based and feature-lifting based accordingly. Result-lifting based depends on domain expertise 
to design 3D representation and resorts to prior knowledge for template matching \cite{mono3d2016chenxiaozhi} or 
geometric constraints for recovery \cite{MousavianAFK17Deep3DBox, li2019gs3d,Zhang_2021_CVPR}. Feature-lifting based 
wants to compute an intermediate representation, say Pseudo LiDAR\cite{wang2019pseudo, XuC18mf3d, Weng_2019_ICCV_Workshops} 
or 3D feature volumes \cite{Chen0SJ20,Guo_2021_ICCV, Reading_2021_CVPR}. \qr{As the absence of depth information, 
we note that works \cite{ChabotCRTC17DeepMANTA, Reading_2021_CVPR, Park_2021_ICCV, Chen0SJ20, Weng_2019_ICCV_Workshops, 
wang2019pseudo, wang2019pseudo++, qian2020e2e, mono3d2016chenxiaozhi} rely on additional data for training.} 
For an autonomous system, redundancy is indispensable to guarantee safety apart from economic concerns,
so image based methods are poised to make a continuing impact over the next few years. 

\subsection{Point Cloud based Methods} \label{subsec:Point-Cloud-based}
Convolutional Neural Networks (CNNs) has always been applied in computer vision for its capability of 
exploiting spatially-local correlations in dense regular grids \cite{lecun2015deep} (\eg, images). \qr{Whereas, point clouds are sparse in 
distribution, irregular in structure, and unordered in geometry by nature. Consequently, convolving against a 
point cloud directly will lead to a gross distortion of representation \cite{li2018pointcnn}. To address issues 
above, works \cite{zhou2018voxelnet, yan2018second, lang2019pointpillars} voxelize their inputs to lend themselves 
to convolution operator adaptively, while works \cite{Yang2020ssd, Yang2020ssd, shi2019pointrcnn} turns to the 
recent advances of learning over point sets \cite{qi2017pointnet, qi2017pointnet++, 3dgnnss, semigcn2017} for help 
instead. Depending on representation learning strategies, we therefore divide existing works into the following 
three groups: (1) voxel based, (2) point based, and (3) voxel-point based. 
}
	
\subsubsection{Voxel based Methods} \label{subsubsec:VoxelbasedMethods}
\qr{
Works in this group voxelize irregular point clouds to 2D/3D compact grids and then collapse it to a 
bird's-eye-view 2D representation on which CNNs effectively convolve against 
\cite{TANet2020,VeloFCN2016li,yang2018pixor,Xu_2021_ICCV,YinCenterPoint,zhou2018voxelnet,
yan2018second,xu2020behind,lang2019pointpillars,ciassd,Zheng_2021_CVPR,voxelrcnn,Sheng_2021_ICCV, Mao_2021_ICCV, 
shi2020points2parts, hvnet2020ye, voxelfpn2020Kuang}.} Typically, VoxelNet \cite{zhou2018voxelnet} 
rasterizes point clouds into volumetric dense grids, followed by 3D CNNs that convolve along each dimension.
Notice that the computation overheads and memory footprints grow cubically in VoxelNet, SECOND \cite{yan2018second} leverages sparse 
convolution operation to get rid of unnecessary computation squandered by unavailing zero-padding voxels. 
To thoroughly remove 3D CNNs layers, PointPillars \cite{lang2019pointpillars} elongates voxels into pillars 
that are arrayed in a BEV perspective. Note that PointPillars decreases the resolution in vertical axis in essence 
which is detrimental to the learning of representation. As a tradeoff, noticeable efforts stretch back to voxels. 
Voxel R-CNN \cite{voxelrcnn} aggregates voxel-wise features from 3D convolutional volumes for proposal refinement.
\qr{SE-SSD \cite{Zheng_2021_CVPR} jointly supervises a student network under the guidance of the distilled knowledge 
injected by its teacher. Works \cite{shi2020points2parts, hvnet2020ye, voxelfpn2020Kuang} incorporate multi-scale 
voxelization strategies into feature aggregation. Works \cite{Sheng_2021_ICCV, Mao_2021_ICCV, TANet2020} exploit 
long-range contextual dependencies among voxels inspired by recent success of Transformers \cite{AttentionisAllyouNeed} 
in computer vision. Works \cite{xu2020behind, Xu_2021_ICCV} integrate shape learning strategy to the deteriorating 
point cloud quality that caused by adverse weather, occlusions, and truncations \etc. It is worth mentioning 
that voxel based approaches benefit from bird-view representation, which possess less scale ambiguity and 
minimal occlusions \cite{VeloFCN2016li,yang2018pixor}.
}
\begin{figure*}[pos=htbp]
  \setlength{\abovecaptionskip}{0.1cm}
  \setlength{\belowcaptionskip}{0cm}
    \centering     
        \includegraphics[width=0.85\textwidth]{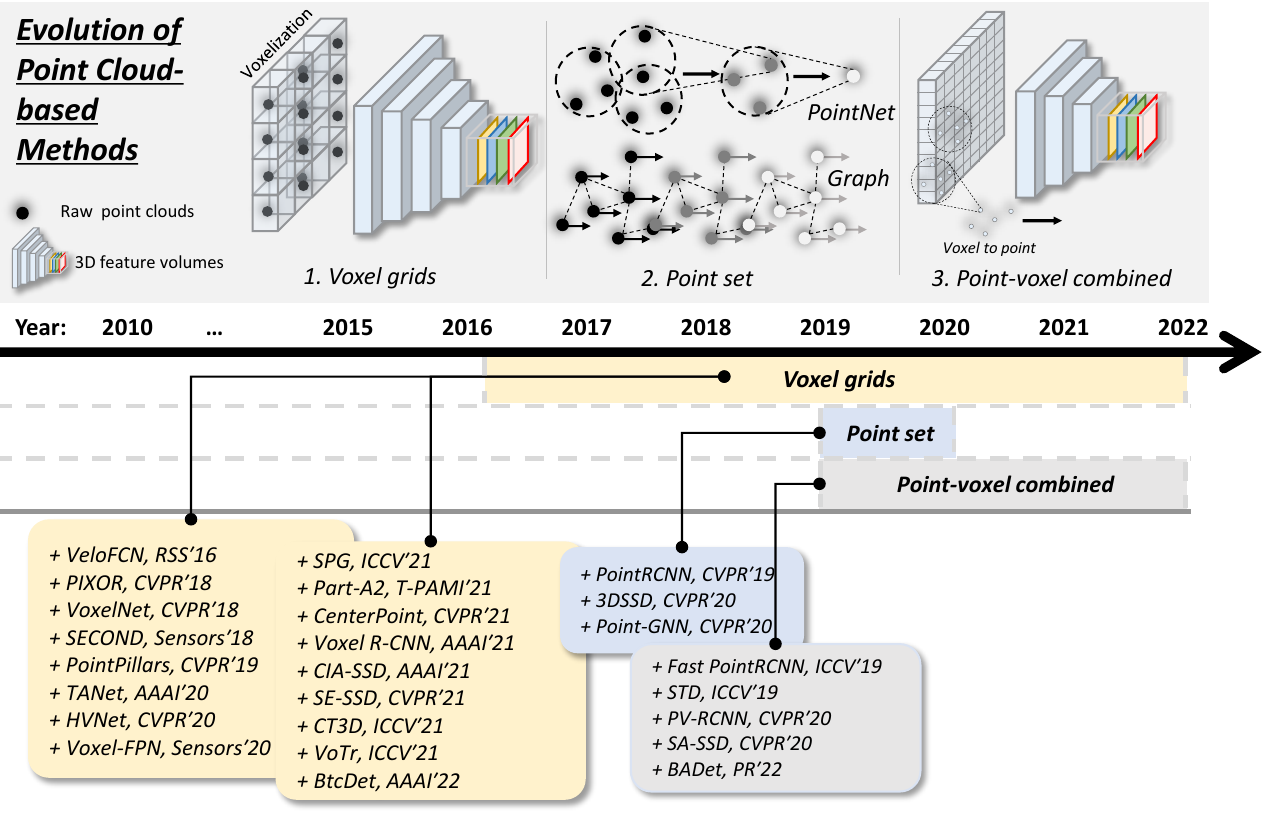} 
        \caption{\textbf{Evolution of point cloud-based methods.}
        }
    \label{subfig:pointcloudbasedmethods} 
\end{figure*}
\subsubsection{Point based Methods} \label{subsubsec:PointbasedMethods}
\qr{
Works in this group utilize permutation invariant operators to implicitly capture local structures 
and fine-grained patterns without any quantization in order to retain the original geometry 
of a raw point cloud \cite{Yang2020ssd, Point-GNN, shi2019pointrcnn}.} To this end, PointRCNN \cite{shi2019pointrcnn} leverages PointNet-like 
block \cite{qi2017pointnet++} to learn semantic cues associated with foreground points over which 3D proposals are 
generated in a bottom-up fashion. Notice that PointNet series are exhausted with the process of upsampling and 
broadcasting semantic cues back into the relevant points, 3DSSD \cite{Yang2020ssd} closely revisits the sampling 
strategy with feature distance, compounded by Euclidean distance, to safely remove it. Encouraged by the promising 
performance on classification and semantic segmentation from point clouds \cite{semigcn2017,graphsage2017,wang2019dynamic}, 
Point-GNN \cite{Point-GNN} reasons on local neighborhood graphs constructed from point clouds, on which each node iteratively 
summarizes semantic cues from intermediate reaches of its neighbors. \qr{Point based approaches are internally 
time-consuming with a ball query complexity of $\mathcal{O}\left( k\cdot \left| \mathcal{X} \right| \right)$. 
Note that works \cite{Yang2020ssd, shi2019pointrcnn} leverage multi-scale and multi-resolution grouping to 
achieve an expanding receptive field, which make latency even severe as points in $\mathcal{X}$ grow. 
}

\subsubsection{Point-voxel based Methods} \label{subsubsec:PointvoxelbasedMethods}
\qr{
Works in this group reveal a new fashion trend to integrate the merits of both worlds together: voxel based 
\cite{zhou2018voxelnet,yan2018second, lang2019pointpillars} approaches are computationally effective but the 
desertion of fine-grained patterns degrades further refinement, while point based methods \cite{Point-GNN, 
Yang2020ssd, shi2019pointrcnn} have relatively higher latency but wholly preserve the irregularity and 
locality of a point cloud \cite{Chen0SJ19, yang2019std, pvrcnn2020, qian_2022_PR,he2020sassd}.} 
STD \cite{yang2019std} applies PointNet++ to summarize semantic cues for sparse points, each of which is 
then voxelized to form a dense representation for refinement. PV-RCNN \cite{pvrcnn2020} 
deeply integrates the effectiveness of 3D sparse convolution \cite{yan2018second} 
and the flexible receptive fields of PointNet-like set abstraction \cite{qi2017pointnet++} to learn more 
discriminative semantics. SA-SSD \cite{he2020sassd} interpolates 3D sparse convolution features 
for raw point clouds on which an auxiliary network is applied to endow voxel features with 
structure-aware capability. \qr{BADet \cite{qian_2022_PR} exploits long-range interactions iteratively among 
detection candidates, wherein local neighborhood graphs are constructed to facilitate a boundary-aware 
receptive field in a coarse-to-fine manner.
}
   
\subsubsection{Summary}
Fig. \ref{subfig:pointcloudbasedmethods} illustrates the evolution of point cloud based. Works in this 
group are divided into voxel based, point based, and point-voxel based respectively, among which voxel based 
appears to be dominant, as evidenced by Fig. \ref{subfig:pointcloudbasedmethods}. Voxel based is easily 
amenable to efficient hardware implementations with distinguished accuracy and relatively lower latency.
Point based easily retains the spatially-local structure of a point cloud at the cost of taking longer 
feedforward time than those based on voxels. \qr{Comparing the three subdivisions, voxel based is still 
the most promising direction currently existing concerning real-time applications, say autonomous driving.
}
\begin{figure*}[pos=htbp]
  \setlength{\abovecaptionskip}{0.1cm}  
  \setlength{\belowcaptionskip}{0cm}    
    \centering        
        \includegraphics[width=0.85\textwidth]{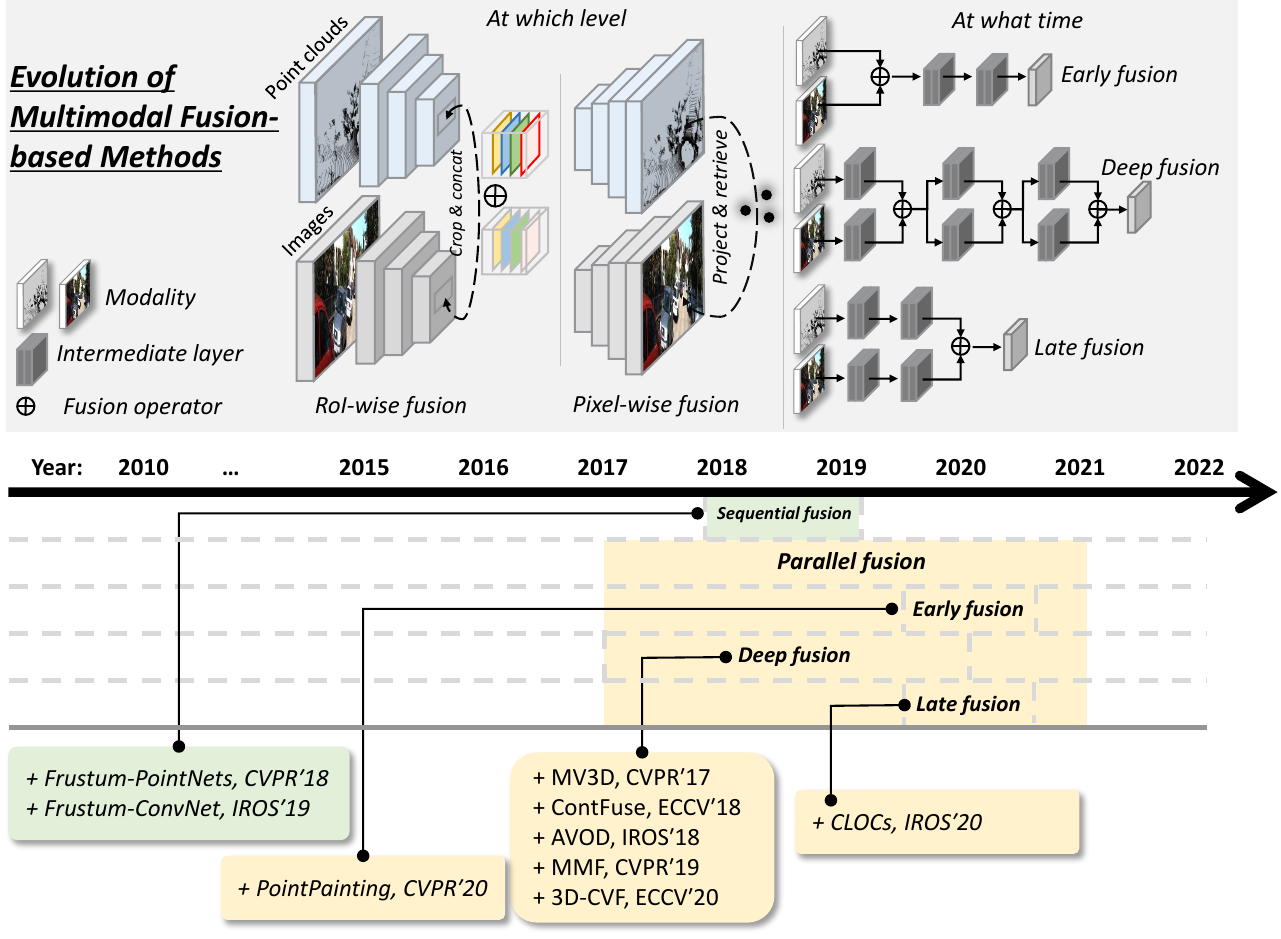} 
        \caption{\textbf{Evolution of multimodal fusion-based methods.}
        }
    \label{subfig:multimodalbasedmethods} 
\end{figure*}

\subsection{Multimodal Fusion based Methods} \label{subsec:Multimodal-Fusion-based}
Intuitively, single modality itself has its own defects, a joint treatment appears to be a potential 
opportunity to failure cases. Nevertheless, multimodal fusion based has still trailed its point cloud based 
counterparts thus far. What, how, and when to fuse have not been full understood yet \cite{feng2020deep}. \qr{Depending 
on to what extent these two modalities are coupled, we divide existing efforts into sequential fusion based and 
parallel fusion based. As the names suggest, if data flow has more than one independent path through the networks, 
then it is considered as \emph{parallel}, otherwise \emph{sequential}. The former emphasizes that the data flow 
of different modalities can pass through networks concurrently, while the latter emphasizes that the data flow of 
different modalities only has one single path to flow through networks successively. In what follows, we analyze 
these two paradigms as regards fusion evolutions, connections, and concerns to meet the requirements.
}

\subsubsection{Sequential Fusion based Methods} \label{subsubsec:SequentialFusionbasedMethods}
\qr{
Works in this group are characterized by 2D driven 3D pipelines in a sequential manner, wherein the input of 
downstream depends heavily on the output of upstream \cite{qi2018frustum, wang2019frustum}. }
Frustum PointNets \cite{qi2018frustum} first leverages a mature 2D CNN object detector to predict 2D proposals, 
each of which are then transformed to 3D space in order to crop corresponding frustum candidates, followed by 
point cloud based detectors for segmentation and detection. Frustum-ConvNet \cite{wang2019frustum} generates a 
sequence of frustums that slide along the optical axis perpendicular to 2D image plane, which are summarized as 
frustum-wise features by PointNet blocks and arrayed as a BEV feature map for fully convolutional network.
Works in this group resort to off-the-shelf 2D detectors for prior knowledge, with 3D search 
space being dramatically reduced. \qr{Whereas, they also indicate a fatal risk that the failure of 
2D detectors is bound to deteriorate all subsequent pipelines, which violates the original intention of 
providing redundancy during difficult conditions rather than just complementary and we therefore 
argue that sequential fusion based may not be well-suited for autonomous driving for the sake of safety. 
}

\subsubsection{Parallel Fusion based methods} \label{subsubsec:ParallelFusionbasedmethods}
\qr{
Works in this group are decoupled from the single modality. Suppose that one of the branches is cut off, 
the network will still work if adjusted appropriately, say hyper-parameters. Depending on at what point 
these semantic representations from single modality are fused, a further subdivision for parallel fusion 
based can be identified: (1) early fusion, (2) deep fusion, and (3) late fusion. 
}

\qr{
\emph{Early fusion} fuses different modalities at the data preprocessing point, which is not commonly used in 
light of the big noise of feature alignment at low level semantics. PointPainting \cite{PointPainting2020Vora} 
takes as inputs raw point clouds, compounded by segmentation scores that are predicted from the associated images, 
with the help of an independent segmentation network. \emph{Deep fusion} fuses different modalities at the 
intermediate point.} MV3D \cite{chen2017multi} takes as inputs the bird-view, front view, and images, fusing 
intermediate convolution features at stage two. AVOD \cite{ku2018avod} further extends 
fusion strategy to stage one to enrich more informative semantics for proposal generation. Observe that works 
\cite{chen2017multi,ku2018avod} fuse futures in a roi-wise manner, ContFuse \cite{ContFuse2018ming} further 
fuses multi-scale convolutional features via continuous convolution in a pixel-wise fashion.
Notice that pixel-wise fusion is vulnerable to faraway objects due to the difficulty of finding corresponding 
pixels for the sparse LiDAR signals at long ranges, MMF \cite{mmf2019cvpr} exploits multiple related subtasks 
(\eg, ground estimation, depth completion, and 2D object detection \etc) to enhance the learning of cross-modality 
fusion. Note that existing efforts treat semantics from different modalities equally, 3D-CVF \cite{3dcvf2020jin}
employs attention mechanism to adaptively fuse semantics from point clouds and the associated images. 
\qr{\emph{Late fusion} fuses the outputs of different modalities at the decision point. CLOCs \cite{CLOCs2020} fuses 
the outputs of 2D and 3D detections in decision level via exploiting the consistency of geometries and semantics.
}
\vspace{1em}
\subsubsection{Summary}
Fig. \ref{subfig:multimodalbasedmethods} illustrates the evolution of multimodal fusion based. \qr{What to 
use. Consider the most commonly used sensors for autonomous driving: camera and LiDAR. Preceding works 
\cite{ qi2018frustum, wang2019frustum, PointPainting2020Vora, chen2017multi, ku2018avod, mmf2019cvpr, 3dcvf2020jin,
CLOCs2020, ContFuse2018ming} all take images and point clouds as inputs. How to fuse. Efforts have been made to 
align semantics of different levels at different scales. Works \cite{chen2017multi, ku2018avod, qi2018frustum, 
wang2019frustum, CLOCs2020} exploit roi-wise fusion. Works \cite{PointPainting2020Vora, mmf2019cvpr, 3dcvf2020jin, 
ContFuse2018ming} leverage point-wise fusion. When to use. Cross-modality fusion can happen at any time 
during the forward propagation: say early fusion \cite{PointPainting2020Vora}, deep fusion \cite{
chen2017multi, ContFuse2018ming, ku2018avod, mmf2019cvpr, 3dcvf2020jin}, and late fusion \cite{CLOCs2020}. 
In retrospect of the evolution of this group, we notice that multimodal fusion based still lags far behind its 
point cloud based counterparts. We attribute the performance gap to the difficulty of semantic alignments. First, semantics 
from images and point clouds are heterogeneous as they are presented in different views \cite{3dcvf2020jin} 
(\ie, camera view vs. real 3D view). Consequently, finding point-wise correspondences is difficult as the 
transformation from LiDAR points to image pixels is a many-to-one mapping and vice versa. Such an ambiguity is 
referred as ``feature blurring'' in PointPainting \cite{PointPainting2020Vora}. Second, images are arrayed in dense 
grids, while point clouds are distributed in sparse points. To what extent semantics are aligned by forcing these 
two modalities to have the same size for aggregation remains untouched. Finally, operations that we use to crop 
and resize features may not be as accurate as expected \cite{CLOCs2020}.
}

\vspace{0.5em}                    
\section{EVALUATION} \label{sec:eval}
 
This section holds an apples-to-apples comparison of the state-of-the-arts on the widely used KITTI dataset, 
more recent nuScenes and Waymo dateset (Sec. \ref{ComprehensiveComparsionoftheState-of-the-Arts}). A case 
study based on fifteen selected models is given in Sec. \ref{CaseStudy}, with respect to runtime analysis 
(Sec. \ref{RuntimeAnalysis}), error analysis (Sec. \ref{ErrorAnalysis}), and robustness analysis (Sec. \ref{RobustnessAnalysis}).

\subsection{Comprehensive Comparison of the State-of-the-Arts} \label{ComprehensiveComparsionoftheState-of-the-Arts}
Table \ref{tab:KITIIdataset} summarizes the 3D detection performance on the KITTI Dataset. 
Image based methods have currently trailed the performance of point clouds counterparts thus far, which 
should be ascribed to depth ambiguity. Point clouds based methods still predominate KITTI benchmark due to 
low latency and high accuracy by resorting to 3D sparse convolution. Multimodal fusion based methods are closing the gap with point clouds counterparts 
somewhat. As mentioned above, fusing these two modalities together is non-trivial due to view misalignment. Noticeably,  
on the one hand, monocular or stereo cameras indeed bring extra source information as a supplementary, 
which circumvent the risk of over-reliance on a single sensor. On the other hand, multiple sensors hinder the 
runtime and make it hard to deploy given the requirement of continuous synchronization. \qr{Table \ref{tab:nuScenesdataset} and 
Table \ref{tab:Waymodataset} present the 3D detection performance on the more recent nuScenes dataset and Waymo dataset, respectively.
Although works that report performances on nuScenes or Waymo are not as common as KITII currently, 
assessing the effectiveness of detectors on these two large-scale datasets shall be necessary in the foreseeable 
future, regardless of dataset size or diversity as evidenced in Table \ref{tab:datasets}.} 

\begin{table*}[htbp]
    \setlength{\abovecaptionskip}{0.1cm}
    \setlength{\belowcaptionskip}{0.2cm}
        \begin{center}
            \caption{\textbf{Comparisons of the state-of-the-art 3D detection $ {AP}_{3D}|_{R_{40}}$ 
            on KITTI \emph{test} split}, by submitting to official test server. All these methods follow the official KITTI evaluation protocol, \ie
            the rotated ${IoU}_{3D}$ of 0.7, 0.5, and 0.5 is for the categories of $Car$, $Cyclist$, and $Pedestrian$, respectively.
            `-' means the results are unavailable.}
                \resizebox{\textwidth}{9.35cm}{
                \begin{tabular}{ccrrp{0.5cm}<{\centering}*3cp{0.25cm}<{\centering}*3cp{0.25cm}<{\centering}p{1.5cm}<{\centering}p{1cm}<{\centering}}
                \Xhline{1pt}
                \multicolumn{1}{c}{\multirow{2}{*}{\makecell[{c}{c}]{\textbf{Input}}}} & \multicolumn{2}{c}{\multirow{2}{*}{\makecell[{}{p{2cm}}]{\textbf{Methods}}}} &  
                \multicolumn{1}{c}{\multirow{2}{*}{\makecell[{c}{c}]{\textbf{Speed}\\ \textbf{(fps)}}}}&
                \multicolumn{3}{c}{\textbf{Cars}} & & \multicolumn{3}{c}{\textbf{Pedestrians}} & & \multicolumn{ 3}{c}{\textbf{Cyclists}} \\
                \cline{5-7}\cline{9-11}\cline{13-15}
                \multicolumn{1}{c}{}& \multicolumn{2}{c}{} & \multicolumn{1}{c}{}& \multicolumn{1}{c}{\makecell[{}{c}]{\textbf{Easy}}} &\multicolumn{1}{c}{\makecell[{}{c}]{\textbf{Moderate}}} & \multicolumn{1}{c}{\makecell[{}{c}]{\textbf{Hard}}} & 
                                                                                                            &  \multicolumn{1}{c}{\makecell[{}{c}]{\textbf{Easy}}} &\multicolumn{1}{c}{\makecell[{}{c}]{\textbf{Moderate}}} & \multicolumn{1}{c}{\makecell[{}{c}]{\textbf{Hard}}}  & 
                                                                                                            &  \multicolumn{1}{c}{\makecell[{}{c}]{\textbf{Easy}}} &\multicolumn{1}{c}{\makecell[{}{c}]{\textbf{Moderate}}} & \multicolumn{1}{c}{\makecell[{}{c}]{\textbf{Hard}}} \\
    \Xhline{0.7pt}
                \multirowcell{20}{Images} & \multirowcell{11}{Result-\\lifting} 
                 &Deep MANTA \cite{ChabotCRTC17DeepMANTA}           & -                       & -   & -   & -                 & &               -      & -      & -                     &  &                   -        & -        & -     \\
                & &Deep3DBox \cite{MousavianAFK17Deep3DBox}           & -                       & -   & -   & -                 & &               -      & -      & -                     &  &                   -        & -        & -     \\
                & &Mono3D \cite{mono3d2016chenxiaozhi}           & -                       & 2.53   & 2.31   &2.31                 & &               -      & -      & -                     &  &                   -        & -        & -     \\
		 & &GS3D \cite{li2019gs3d}                &  -                      & 4.47   & 2.90   & 2.47                & &                   -    &    -   &     -               &  &                         -    &   -     &  -     \\
                & &Mono3D-PLiDAR \cite{Weng_2019_ICCV_Workshops} & -                       & 1.76    & 7.50    & 6.10                  & &                  -    &    -   &  -                   &  &                       -     &  -      &   -   \\
		 & &Monodle \cite{Ma_2021_CVPR}   &  25                      & 17.23 &12.26&10.29                & &                 9.64 &6.55 &5.44                 &  &                      4.59 &2.66 &	2.45      \\
		 & &MonoRCNN \cite{Shi_2021_ICCV}   &  -                      & 18.36 &12.65  &10.03                & &                  -      &   -    &  -                 &  &                       -      &  -       & -      \\
		 & &MonoFlex \cite{Zhang_2021_CVPR}   &  -                      & 19.94 &	13.89 &12.07                & &                 9.43 &6.31 &5.26                 &  &              4.17 &2.35 &	2.04      \\
		 & &DD3D \cite{Park_2021_ICCV}   &  -                      & 23.19 &16.87 & 14.36                & &                 16.64   & 11.04 &	9.38                 &  &                      7.52 & 4.79 &	4.22      \\
		 & &Stereo R-CNN \cite{li2019stereo}   &  -                      & 47.58 &30.23  &23.72                & &                  -      &   -    &  -                 &  &                       -      &  -       & -      \\
		 & &YOLOStereo3D \cite{2021YOLOStereo3D}   &  10                      & 65.68 &41.25 & 30.42                & &                28.49 &19.75 &16.48                 &  &              - &- &	-      \\
     \cline{2-15}
                & \multirowcell{10}{Feature-\\lifting} 
                 &  3DOP \cite{3dop2018chen,chen2015nips3dop}               &   -                     & -       & -       & -                     & &                  -      &   -    &  -                 &  &                       -      &  -       & -     \\
                 & & MF3D \cite{XuC18mf3d}           & -                       & -   & -   & -                 & &               -      & -      & -                     &  &                   -        & -        & -     \\  
                 & & OFT-Net\cite{OFT-NetBMVC}           & -                       & 2.50   & 3.28   & 2.27                 & &               -      & -      & -                     &  &                   -        & -        & -     \\  
                & & Mono3D-PLiDAR \cite{Weng_2019_ICCV_Workshops}           & -                       & 10.76 &7.50 &6.10                  & &               -      & -      & -                     &  &                   -        & -        & -     \\  
                & &CaDDN \cite{Reading_2021_CVPR}  &-                    &19.17 &13.41 &11.46                 & &                       12.87 &	8.14 & 6.76            &  &                       7.00 &3.41 &3.30     \\    
		 & &Pseudo-LiDAR \cite{wang2019pseudo}         &-                    &54.53  &34.05  &28.25                & &                        -   &  -     &    -             &  &                        -     &  -       &   -     \\
                & &Pseudo-LiDAR++ \cite{wang2019pseudo++}     &-                    &61.11    &42.43  &36.99                & &                        -   &  -     &    -             &  &                        -     &  -       &   -     \\
                & &Pseudo-LiDAR E2E \cite{qian2020e2e}  &-                    &64.75   &43.92 &38.14                 & &                        -   &  -     &    -             &  &                        -     &  -       &   -     \\    
		  & &DSGN \cite{Chen0SJ20}         &-                    &73.50 &52.18 &	45.14                & &       20.53 & 15.55 & 14.15                &  &        27.76 &	18.17 & 16.21   \\
                & &LIGA-Stereo \cite{Guo_2021_ICCV}     &-                    &81.39 &64.66 &57.22                & &                        40.46& 30.00 & 27.07             &  &                       54.44 & 36.86 &	32.06    \\
    \hline
                \multirowcell{25}{Point clouds} & \multirowcell{17}{Voxel\\based}  
		     & VeloFCN \cite{VeloFCN2016li}           &  1.0                     &   -    &   -     &    -                   & &                      -   &  -     &  -                &  &                        -     &  -       &  -    \\
                & &PIXOR \cite{yang2018pixor}               &  28.6                   &    -    &     -  &   -                   & &                       -  &   -    &  -                 &  &                        -     &  -       &  -    \\
		 & & HVNet \cite{hvnet2020ye}               &   31                    &    -    &     -   &   -                   & &                         -  &   -    &  -                &  &                        -     &  -       &  -    \\
                & & VoxelNet\cite{zhou2018voxelnet}           & 2.0      & 77.82  & 64.17 & 57.51               & &                       -   &  -     &  -                &  &                         -     &  -       &  -\\                
		 &  &CenterPoint \cite{YinCenterPoint}               &  -                   &  81.17 &	73.96 & 69.48                 & &                       47.25 & 39.28 & 36.78                 &  &             73.04 & 56.67 & 50.60    \\                
		 & &PointPillars \cite{lang2019pointpillars}         & 62.0                   &  82.58 & 74.31 & 68.99               & &                 51.45   & 41.92 & 38.89           & &                        77.10  & 58.65 & 51.92 \\
                &    &TANet \cite{TANet2020}           & -                      &   84.39 &75.94 & 68.82                  & &                     53.72 &	44.34 &	40.49                &  &                75.70 &	59.44 &	52.53   \\
                & &SECOND \cite{yan2018second}            &  26.3                  &  84.65 & 75.96 & 68.71               & &                       -   &  -     &  -                &  &                         -     &  -       &  -\\
		 & &Voxel-FPN \cite{voxelfpn2020Kuang}         &   50                    & 85.48 & 76.70 & 69.44               & &                        -   &  -     &    -             &  &                        -     &  -       &   -     \\
		 & &Part-A$^2$ \cite{shi2020points2parts}        &  12.5                   &  87.81 & 78.49 & 73.51                & &                 53.10   & 43.35 & 40.06          &  &                       79.17  & 63.52 & 56.93 \\                                        
		 & &CIA-SSD \cite{ciassd}         & 32                   &  89.59 & 80.28 & 72.87               & &                 -   & - & -           & &                        -  & - & - \\
		 & &Voxel R-CNN \cite{voxelrcnn}         & 25.2                   & 90.90 &	81.62 & 77.06               & &                 -   & - & -           & &                        -  & - & - \\
		 & &CT3D \cite{Sheng_2021_ICCV}         & -                   & 87.83 & 81.77 & 77.16               & &                 -   & - & -           & &                        -  & - & - \\
		 & &VoTr \cite{Mao_2021_ICCV}         & -                   & 89.90 & 82.09 & 79.14               & &                 -   & - & -           & &                        -  & - & - \\
		 & &SPG \cite{Xu_2021_ICCV}               &  -                   &  90.50 &82.13 &	78.90                 & &                       -  &   -    &  -                 &  &                        -     &  -       &  -    \\
		 & &SE-SSD \cite{Zheng_2021_CVPR}         &32                   & 91.49 &	82.54 & 77.15               & &                 -   & - & -           & &                        -  & - & - \\
		 & &BtcDet \cite{xu2020behind}               &  -                   &  90.64 & 82.86 & 78.09                 & &                       -  &   -    &  -                 &  &                       82.81 &68.68 &61.81    \\
    \cline{2-15}
                & \multirowcell{3}{Point\\based} 
                    &PointRCNN \cite{shi2019pointrcnn}       &   10                    &  86.96 & 75.64& 70.70                & &                    47.98 & 39.37& 36.01          &  &                         74.96 & 58.82& 52.53 \\
		  & &Point-GNN \cite{Point-GNN}        & 1.7                    & 88.33   & 79.47& 72.29                & &                      51.92 & 43.77& 40.14         & &                          78.60 & 63.48 & 57.08 \\        
                & &3DSSD \cite{Yang2020ssd}              &25.0                  & 88.36  & 79.57& 74.55                 & &                   54.64  &44.27 &40.23         &  &                          82.48 & 64.10 & 56.90 \\
    \cline{2-15}
                & \multirowcell{5}{Point-voxel\\based} 
                    &Fast PointRCNN \cite{Chen0SJ19} & 16.7                  & 85.29 & 77.40 &70.24                  & &                     -        &  -      & -              & &                         -       & -       & -        \\
                & &STD \cite{yang2019std}                    & 12.5                   &87.95  & 79.71 & 75.09                  & &                     53.29  & 42.47 & 38.35        & &                          78.69 &61.59 & 55.30  \\  
                & &SA-SSD \cite{he2020sassd}              & 25.0                  &88.75  & 79.79 & 74.16                  & &                        -   &  -     &    -             &  &                             -     &  -       &   -     \\                       
                & &PV-RCNN \cite{pvrcnn2020}           & 12.5                   &90.25  & 81.43 & 76.82                  & &                       52.17 &43.29  &40.29         & &                          78.60 &63.71 & 57.65        \\
                & &BADet \cite{qian_2022_PR}           & 7.1                   &89.28 &	81.61 &76.58                   & &                       -        &  -      & -              & &                         -       & -       & -         \\
    \hline
                \multirowcell{10}{Multimodal} & \multirowcell{2}{Sequential \\fusion} 
                    &Frustum-PointNets \cite{qi2018frustum}   &5.9                 & 82.19   & 69.79  & 60.59               & &                   50.53  & 42.15 &38.08          & &                               72.27 & 56.12 & 49.01 \\
                & &Frustum-Convnet \cite{wang2019frustum}            &2.1                  & 87.36  &76.39   &66.69                & &                      52.16 & 43.38 & 38.80        & &                                 81.98 &65.07 &56.54 \\
                                
    \cline{2-15}
                & \multirowcell{7}{Parallel \\fusion}
	          &MV3D \cite{chen2017multi}                  & 2.8                 & 74.97  & 63.63  & 54.00               & &                        -   &  -     &    -             &  &                              -     &  -       &   -     \\
                 & &AVOD \cite{ku2018avod}                  &12.5                & 76.39 & 66.47   & 60.23               & &                    36.10  & 27.86  &25.76         & &                               57.19 & 42.08& 38.29\\
                & &ContFuse \cite{ContFuse2018ming}             & 16.7               & 83.68 & 68.78   & 61.67                & &                         -   &  -     &    -             &  &                               -     &  -       &   -     \\
                &    &PointPainting \cite{PointPainting2020Vora}        &2.5       & 82.11   &71.70     &67.08               & &                      50.32 & 40.97 &37.87         & &                                77.63 &63.78  &55.89\\
                & &MMF \cite{mmf2019cvpr}                   & 12.5               & 88.40 & 77.43   & 70.22               & &                         -   &  -     &    -             &  &                                -     &  -       &   -     \\
                & &3D-CVF \cite{3dcvf2020jin}               & -                   & 89.20  &80.05 &73.11                  & &                        -   &  -     &    -             &  &                        -     &  -       &   -     \\                    
                & &CLOCs \cite{CLOCs2020}               & -                   & 88.94 & 80.67 & 77.15                 & &                        -   &  -     &    -             &  &                        -     &  -       &   -     \\                    
        \Xhline{1pt}
        \end{tabular}}
        \label{tab:KITIIdataset}
        \end{center}
        \vspace{-1.5em}
\end{table*}
\begin{table*}[htbp] 
    \centering
    \renewcommand\arraystretch{1.2}
    \caption{\textbf{Comparisons of the state-of-the-art 3D detection on nuScenes \emph{test} set.} 
     ``TC'' and ``Cons.Veh.'' denote traffic cone and construction vehicle respectively.}
        \newcommand{\tabincell}[2]{\begin{tabular}{@{}#1@{}}#2\end{tabular}} 
        \resizebox{\textwidth}{!}{
        \begin{tabular}{rccccccccccccccc}            
        \toprule
        \textbf{Methods} &   \textbf{mAP} &  \textbf{NDS} &  \textbf{Car} & \textbf{Truck} &\textbf{Bus} &\textbf{Trailer} &\textbf{CV} &\textbf{Pedestrian} &\textbf{Motorcycle} &\textbf{Bicycle} &\textbf{TC} &\textbf{Barrier} \\
    \cmidrule(r){1-1}\cmidrule(lr){2-3}\cmidrule(l){4-13}
    PointPillars \cite{lang2019pointpillars} &30.5 & 45.3& 68.4 &23.0 & 28.2 & 23.4 & 4.1 & 59.7 & 27.4 & 1.1 & 30.8 & 38.9 \\
    PointPainting \cite{PointPainting2020Vora} & 46.4 & 58.1&77.9&35.8&36.2&37.3&15.8&73.3&41.5&24.1&62.4&60.2\\
    CenterPoint \cite{YinCenterPoint} &58.0&65.5&84.6&51.0&60.2&53.2&17.5&83.4&53.7&28.7&76.7&70.9\\
        \bottomrule									
    \end{tabular}%
    \label{tab:nuScenesdataset}
    }   
\end{table*}%
\begin{table*}[htbp] 
    \centering
    \renewcommand\arraystretch{1.2}
    \caption{\textbf{Comparisons of the state-of-the-art 3D detection on Waymo \emph{val} set.}}
        \newcommand{\tabincell}[2]{\begin{tabular}{@{}#1@{}}#2\end{tabular}} 
        \resizebox{\textwidth}{!}{
        \begin{tabular}{rccp{0.2\textwidth}<{\centering}p{0.2\textwidth}<{\centering}p{0.2\textwidth}<{\centering}}
        \toprule
        \multirow{2}*{\textbf{Methods}} &\multirow{2}*{ \tabincell{c}{\textbf{LEVEL$\_$1} \\ \textbf{3D mAP/mAPH}}} &\multirow{2}*{ \tabincell{c}{\textbf{LEVEL$\_$2} \\\textbf{3D mAP/mAPH}}} & \multicolumn{3}{c}{\textbf{LEVEL$\_$1 3D mAP/mAPH by Distance}} \\
        & & & \textbf{0-30m} & \textbf{30-50m} & \textbf{50m-Inf} \\
        \midrule
        PointPillars \cite{lang2019pointpillars}  & 63.30/62.70 & 55.20/54.70 & 84.90/84.40 & 59.20/58.60 & 35.80/35.20\\
        Voxel R-CNN \cite{voxelrcnn}  & 75.59/- & 66.59/- & 92.49/- & 74.09/- & 53.15/-\\
        SECOND \cite{yan2018second}  & 67.94/67.28 & 59.46/58.88 & 88.10/87.46 & 65.31/64.61 & 40.36/39.57\\
        PV-RCNN \cite{pvrcnn2020}  & 71.69/71.16 & 64.21/63.70 &91.83/91.37 & 69.99/69.37 & 46.26/45.41\\
        VoTr \cite{Mao_2021_ICCV}  & 74.95/74.25 & 65.91/65.29 & 92.28/91.73 & 73.36/72.56 & 51.09/50.01\\
        BtcDet \cite{xu2020behind} & 78.58/78.06 & 70.10/69.61 & 96.11/- & 77.64/- & 54.45/-\\
        \bottomrule									
    \end{tabular}%
    \label{tab:Waymodataset}
    }
\end{table*}%

\subsection{Case Study} \label{CaseStudy}
\vspace{0.5em}
\subsubsection{Experimental Setup}
Fifteen models are selected from the surveyed works depending on whether the official source code and the 
corresponding pretrained parameters are available. Notice that all our experiments are conducted on KITTI dataset 
by directly reloading official pretrained parameters with default settings. We follow KITII protocol to evaluate 
on the \emph{val} split for the most important \emph{Car} category of moderate difficulty based on $AP_{3D}|_{R_{11}}$ 
metric with an IoU threshold 0.7. For the ease of reproducibility, all materials are available online:
\href{https://github.com/rui-qian/SoTA-3D-Object-Detection}{\emph{https://github.com/rui-qian/SoTA-3D-Object-Detection}}.
\vspace{0.8em}
\subsubsection{Runtime Analysis} \label{RuntimeAnalysis} 
\vspace{0.3em}
\qr{
To assess the real latency of detectors, we report runtime analysis in Table \ref{tab:RuntimeAnalysis}. 
Instead of just citing the numbers claimed in the papers, we conduct new experiments by ourselves. We argue that it is 
necessary as these numbers are obtained under different hardware resources in various settings. 
Some of them may use Tesla P40 GPU (\eg, Fast PointRCNN \cite{Chen0SJ19}) whereas others may use TITAN Xp GPU 
(\eg, CIA-SSD \cite{ciassd}). Some of them may ignore the time of data processing while others may use multiple process. 
Hence, directly comparing against these numbers inevitably leads to controversy. By contrast, we report two versions 
of runtime on a single GTX 1080Ti GPU, termed as FPS$_{default}$ and FPS$_{unified}$. FPS$_{default}$ means the numbers 
are obtained with official settings, which are almost consistent with the ones claimed in the papers. FPS$_{default}$ reveals 
that our environment can reproduce the claimed results. FPS$_{unified}$ justifies all models in a unified criteria 
with 1 batch size and 4 multiple processes, fully eliminating other irrelevant factors. Table \ref{tab:RuntimeAnalysis} 
indicates CIA-SSD \cite{ciassd} and SE-SSD \cite{Zheng_2021_CVPR} drop a lot under our paradigm. Considering that FLOPS 
is hardware independent, we also provide their FLOPS in the 2$^{nd}$ row that have never been reported before among existing 
surveys to the best of our knowledge. SE-SSD \cite{Zheng_2021_CVPR} shows a superior performance of speed-accuracy tradeoff, 
as evidenced by the 1$^{st}$ and 2$^{nd}$ rows of Table \ref{tab:RuntimeAnalysis}.}

\begin{table*}[htbp]  \scriptsize
    \begin{adjustwidth}{-4in}{-4in}
    \setlength{\tabcolsep}{1pt}  
    \centering
    \renewcommand\arraystretch{1.4}
    \caption{\textbf{Runtime Analysis.} Numbers in 1$^{st}$ row indicate 3D detection on moderate difficulty for \emph{Car} category sorted in ascending order by $AP_{3D}|_{R_{11}}$.}
    \newcommand{\tabincell}[2]{\begin{tabular}{@{}#1@{}}#2\end{tabular}} 
    \resizebox{1.17\textwidth}{!}{
      \begin{tabular}{lccccccccccccccc}
      \toprule
      \textbf{Methods} & \textbf{CaDDN}\cite{Reading_2021_CVPR} & \textbf{PointPillars}\cite{lang2019pointpillars} & \textbf{TANet}\cite{TANet2020}  & \textbf{Point-GNN}\cite{Point-GNN} & \textbf{SECOND}\cite{yan2018second} & \textbf{PointRCNN}\cite{Chen0SJ19} & \textbf{3DSSD}\cite{Yang2020ssd} & \textbf{Part-A$^2$}\cite{shi2020points2parts} & \textbf{SA-SSD}\cite{he2020sassd}  & \textbf{CIA-SSD}\cite{ciassd} & \textbf{PV-RCNN}\cite{pvrcnn2020} & \textbf{Voxel R-CNN}\cite{voxelrcnn} & \textbf{CT3D}\cite{Sheng_2021_ICCV} & \textbf{BADet}\cite{qian_2022_PR} & \textbf{SE-SSD}\cite{Zheng_2021_CVPR} \\
      \midrule
      AP    & 19.19 & 77.28 & 77.54 & 78.34 & 78.62 & 77.38 & 79.23 & 79.38 & 79.80  & 79.74 & 83.56 & 84.54 & 85.47 & 86.21 & 86.25 \\
      FLOPS & 27.33G & 3.72G & 13.54G & 4.45M & 3.49G & 1.53G & 31.44G & 2.95G & 155.48G & 24.48G & 2.97G & 0.87G & 3.17G & 158.74G & 24.45G \\
      Params & 67.55M & 4.83M & 6.5M & 1.49M & 5.33M & 4.04M & 7.56M & 63.81M & 5.34M & 3.81M & 13.12M & 7.59M & 7.83M & 5.79M & 3.81M \\
      FPS$_{default}$ & 3.17  & 23.85 & 14.36 & 1.89  & 15.00    & 6.25  & 10.03 & 9.51  & 20.50  & 33.40  & 7.19  & 16.97 & 7.13  & 7.10   & 32.40 \\
      FPS$_{unified}$ & 3.17  & 23.98 & 14.36 & - & 15.10  & 6.27  & 10.09 & 9.47  & 20.40  & 23.90  & 7.20   & 17.28 & 8.21  & 7.20   & 23.70 \\
      \bottomrule
      \end{tabular}%
    \label{tab:RuntimeAnalysis}%
    }
\end{adjustwidth}
\end{table*}%

\begin{table*}[htbp] 
    \begin{adjustwidth}{-4in}{-4in}
    \setlength{\tabcolsep}{1pt}
    \centering
    \renewcommand\arraystretch{1.4}
    \caption{\textbf{Error Analysis.} We replace the predicted 3D bounding boxes partially with their corresponding 
    ground truth values. Numbers in 1$^{st}$ row indicate 3D detection on moderate difficulty for \emph{Car} category sorted in ascending order by $AP_{3D}|_{R_{11}}$.}
    \newcommand{\tabincell}[2]{\begin{tabular}{@{}#1@{}}#2\end{tabular}} 
    \resizebox{1.17\textwidth}{!}{
      \begin{tabular}{lccccccccccccccc}
      \toprule
      \textbf{Methods} & \textbf{CaDDN}\cite{Reading_2021_CVPR} & \textbf{PointPillars}\cite{lang2019pointpillars} & \textbf{TANet}\cite{TANet2020}  & \textbf{Point-GNN}\cite{Point-GNN} & \textbf{SECOND}\cite{yan2018second} & \textbf{PointRCNN}\cite{Chen0SJ19} & \textbf{3DSSD}\cite{Yang2020ssd} & \textbf{Part-A$^2$}\cite{shi2020points2parts} & \textbf{SA-SSD}\cite{he2020sassd}  & \textbf{CIA-SSD}\cite{ciassd} & \textbf{PV-RCNN}\cite{pvrcnn2020} & \textbf{Voxel R-CNN}\cite{voxelrcnn} & \textbf{CT3D}\cite{Sheng_2021_ICCV} & \textbf{BADet}\cite{qian_2022_PR} & \textbf{SE-SSD}\cite{Zheng_2021_CVPR} \\
      \midrule
      baseline & 19.19 & 77.28 & 77.54 & 78.34 & 78.62 & 77.38 & 79.23 & 79.38 & 79.80  & 79.74 & 83.56 & 84.54 & 85.47 & 86.21 & 86.25 \\
      w/ gt 3D location & 32.20  & 87.20  & 87.08 & 88.79 & 88.31 & 87.31 & 88.72 & 88.11 & 89.45 & 89.15 & 88.40  & 88.53 & 88.65 & 89.06 & 89.09 \\
      w/ gt depth & 30.32 & 78.68 & 78.66 & 79.02 & 82.59 & 78.08 & 85.97 & 83.47 & 86.88 & 86.58 & 84.18 & 84.80  & 86.19 & 86.70  & 86.56 \\
      w/ gt 3D size & 19.36 & 78.42 & 77.88 & 78.80  & 79.16 & 78.07 & 85.69 & 83.36 & 86.26 & 79.80  & 83.95 & 84.47 & 85.86 & 86.34 & 86.25 \\
      w/ gt orientation & 19.38 & 77.52 & 77.73 & 78.46 & 78.72 & 77.56 & 79.24 & 79.42 & 86.23 & 79.78 & 83.68 & 84.32 & 85.58 & 86.28 & 86.33 \\
      \bottomrule
      \end{tabular}%
    \label{tab:ErrorAnalysis}%
    }
\end{adjustwidth}
\end{table*}%

\begin{table*}[htbp] 
    \begin{adjustwidth}{-4in}{-4in}
    \setlength{\tabcolsep}{1pt}
    \centering
    \renewcommand\arraystretch{1.4}
    \caption{\textbf{Robustness Analysis.} Numbers in 1$^{st}$ row indicate 3D detection on moderate difficulty for \emph{Car} category sorted in ascending order by $AP_{3D}|_{R_{11}}$.}
    \newcommand{\tabincell}[2]{\begin{tabular}{@{}#1@{}}#2\end{tabular}} 
    \resizebox{1.17\textwidth}{!}{
      \begin{tabular}{lccccccccccccccc}
      \toprule
      \textbf{LiDAR beams} & \textbf{CaDDN}\cite{Reading_2021_CVPR} & \textbf{PointPillars}\cite{lang2019pointpillars} & \textbf{TANet}\cite{TANet2020}  & \textbf{Point-GNN}\cite{Point-GNN} & \textbf{SECOND}\cite{yan2018second} & \textbf{PointRCNN}\cite{Chen0SJ19} & \textbf{3DSSD}\cite{Yang2020ssd} & \textbf{Part-A$^2$}\cite{shi2020points2parts} & \textbf{SA-SSD}\cite{he2020sassd}  & \textbf{CIA-SSD}\cite{ciassd} & \textbf{PV-RCNN}\cite{pvrcnn2020} & \textbf{Voxel R-CNN}\cite{voxelrcnn} & \textbf{CT3D}\cite{Sheng_2021_ICCV} & \textbf{BADet}\cite{qian_2022_PR} & \textbf{SE-SSD}\cite{Zheng_2021_CVPR} \\
      \midrule
      64 (Baseline) & n.a.  & 77.28 & 77.54 & 78.34 & 78.62 & 77.38 & 79.23 & 79.38 & 79.80  & 79.74 & 83.56 & 84.54 & 85.47 & 86.21 & 86.25 \\
      32    & n.a.  & 68.32 & 67.09 & 75.85 & 75.58 & 70.11 & 77.27 & 76.06 & 76.10  & 76.32 & 77.60  & 76.43 & 78.15 & 76.47 & 76.11 \\
      16    & n.a.  & 57.26 & 46.23 & 58.75 & 59.09 & 57.06 & 59.65 & 59.41 & 57.06 & 56.22 & 63.34 & 58.61 & 60.82 & 57.39 & 57.27 \\
      8     & n.a.  & 32.49 & 24.25 & 32.56 & 31.58 & 34.26 & 31.53 & 34.22 & 30.55 & 29.50  & 35.80  & 34.05 & 38.96 & 30.29 & 31.11 \\
      \bottomrule
      \end{tabular}%
    \label{tab:RobustnessAnalysis}%
    }
\end{adjustwidth}
\vspace{1em}
\end{table*}%
     
\subsubsection{Error Analysis} \label{ErrorAnalysis}
\qr{
To identify the key parameters affecting the performance of detectors, we report error analysis in Table 
\ref{tab:ErrorAnalysis}. As mentioned in Sec. \ref{Foundations},  we adopt 7 parameters for an oriented 
3D bounding box. These 7 parameters are treated equally when we regress variables. However, what mainly 
restricts 3D detection performance remains unexplored largely. Inspired by Monodle \cite{Ma_2021_CVPR}, we 
therefore conduct an errors analysis by replacing part of predicted 3D bounding box parameters with their 
corresponding ground truth values. A prediction will be assigned with a ground truth if the ratio of overlapping 
area exceeds a certain threshold. We set 0.6 in this paper. As shown in Table \ref{tab:ErrorAnalysis}, we achieve a 
significant $AP_{3D}$ gain among all selected models in the 2$^{nd}$ row if the predicted 3D location is replaced 
by ground truth, where the maximum gain is 13.19\%. According to Table \ref{tab:ErrorAnalysis}, we observe that 
3D location error plays the leading role of error contribution, followed by depth and 3D size error.}

\subsubsection{Robustness Analysis} \label{RobustnessAnalysis}
\qr{
To understand to what extent detectors are resilient to LiDAR sparsity, we report robustness analysis in Table 
\ref{tab:RobustnessAnalysis}. As mentioned in Sec. \ref{Sensors}, LiDAR, typically an HDL-64E Velodyne LiDAR, is 
several orders of magnitude more expensive than camera, which leads to an exorbitant cost for deployment. 
Therefore, resorting to a less dense point cloud for 3D detection is encouraging. In this paper, we use algorithms 
proposed in Pseudo-LiDAR++ \cite{wang2019pseudo++} to sparsify KITII LiDAR signals from 64 to 32, 16, 8 accordingly. 
As shown in Table \ref{tab:RobustnessAnalysis}, Point-GNN \cite{Point-GNN}, SECOND \cite{yan2018second}, 3DSSD \cite{Yang2020ssd} 
Part-A$^2$ \cite{shi2020points2parts}, SA-SSD \cite{he2020sassd} and CIA-SSD \cite{ciassd} maintain a reasonable accuracy 
when LiDAR signals are reduced from 64 to 32. We also observe an obvious performance drop among all models when LiDAR signals 
are reduced from 32 to 16.}   

\vspace{1.3em}
\section{RETROSPECT AND PROSPECT} \label{sec:con}
\vspace{0.3em}
\subsection{Concluding Remarks}

\qr{ This research presents a survey on  3D object detection in the context of autonomous driving, for the 
sake of holding potential interest and relevance for 3D visual data analysis, and consequently facilitating a 
mature taxonomy for the interested audience to either form a structured picture quickly or start their own 
research from scratch easily.
}

\qr{
Depending on what modalities are actually fed into networks during inference, we structure existing literature along 
three dimensions: (1) image based, (2) point cloud based, and (3) multimodal fusion based, allowing us to clarify 
the key challenge that stems from nature properties of modality itself. We attribute challenges to visual appearance 
recovery in the absence of depth information from images, representation learning from partially occluded unstructured 
point clouds, and semantic alignments over heterogeneous features from cross modalities. Having taken a glimpse of evolution 
of 3D object detection, an apples-to-apples comparison of the state-of-the-arts is presented. We notice that 
there is a growing tendency for point clouds based  methods to further broaden accuracy advantages over their image based 
counterparts. A case study on the basis of fifteen selected models is conducted to justify the state-of-the-arts, in 
terms of runtime analysis, error analysis, and robustness analysis. We observe that what mainly restricts the performance
of detection is 3D location error.
}

\qr{
In retrospect of what has been done, we draw concluding remarks for the surveyed works. \emph{Seminal works are profound.} 
VoxelNet \cite{zhou2018voxelnet} takes the first lead to propose an end-to-end trainable network via learning an informative 
3D volumetric representation instead of manual feature engineering as most previous works do. Subsequently SECOND \cite{yan2018second}
exploits sparse convolution operation to only convolve against non-empty voxels, mitigating unnecessary computation 
incurred by unavailing zero-padding voxels in light of the sparsity of point clouds. VoxelNet-like pipelines incorporated with 
sparse convolution have been continuously inspiring their successors ever since. \emph{Auxiliary network learning is 
artful.} SA-SSD \cite{he2020sassd} explores an auxiliary network to endow voxel features with structure-aware capability. 
SE-SSD \cite{Zheng_2021_CVPR} and LIGA-Stereo \cite{Guo_2021_ICCV} exploit distilled intermediate representation and beyond 
from a teacher network, which subtly differs from multi-task learning \cite{mmf2019cvpr} as they are detachable in the 
phase of inference artfully. \emph{Transformers are promising.} A new paradigm of applying transformers \cite{AttentionisAllyouNeed} 
to object detection has recently evolved as the effectiveness of acquiring long-range interactions for faraway objects 
and learning spatial context-aware dependencies for false negatives, wherein CT3D \cite{Sheng_2021_ICCV} and VoTr \cite{Mao_2021_ICCV} 
have achieved a remarkable performance gain. 
}
\subsection{Reflections on Future Work}

\qr{In prospect of what remains to be done, we identify the avenues for future work. 
\emph{Safety is nothing without security.} Reasoning under uncertainty matters. 3D visual data holds potentials of 
uncertainty by nature, regardless of LiDAR signals or images. Therefore, all critical decisions that autonomous driving 
system makes should be under the guidance of uncertainty constraints, \eg the system needs to recognize a higher 
uncertainty in foggy weather than that under sunny conditions. Whereas, how to trade off the concerns between
``minimizing risks'' and ``completing tasks'' is still an imperative yet largely unexplored problem \cite{feng2020deep}. 
We notice MonoFlex \cite{Zhang_2021_CVPR} and MonoRCNN \cite{Shi_2021_ICCV} take certainty into account in their works 
among our surveyed literature. Adversarial attack matters. Note that modern autonomous driving system relies heavily 
on deep learning. Whereas, deep learning methods have already been proved to be vulnerable to visually imperceptible 
perturbations \cite{Adversarial2020Tu} and therefore poses an inherent security risk. With sabotage and threats of blind 
spots on the rise, adversarial attacks on 3D object detection should arouse enough attention of 3D vision community.
\emph{Rethink what we have on hand.} Representation matters. Whether the data representation or the discrepancy in 
depth estimation mainly results in the performance gap between images and LiDAR signals remains open. Pseudo-LiDAR 
series \cite{wang2019pseudo,wang2019pseudo++} break the stereotype that images can only be leveraged in the form of 2D 
representation, which remind us to rethink the off-the-shelf 3D sensor data. DSGN \cite{Chen0SJ20} and CaDDN \cite{Reading_2021_CVPR} 
try to learn a latent depth distribution for an intermediate 3D representation directly instead of resorting to 
empirical geometric constraints. These findings also provide a new idea for multimodal fusion as they remove the 
requirement of point-wise correspondences retrieval for semantic alignments. Shape learning matters. A point 
cloud is self-occluded by nature, which makes itself 2.5D in practice. Shape learning from partially occluded 
sparse point clouds seems to be necessary, as evidenced in SE-SSD \cite{Zheng_2021_CVPR}, SPG \cite{Xu_2021_ICCV}, 
and BtcDet \cite{xu2020behind}. In summary, we hope that this survey will shed light on 3D object detection and 
inspire more follow-up literature in this area.
}

\medskip

\textbf{Acknowledgements}. This work was supported by the National Natural Science Foundation of China 
under Grant 62172420 and Beijing Natural Science Foundation under Grant 4202033.

\bibliographystyle{elsarticle-num}
   
\vspace{1.8em}
\bibliography{references}

\end{document}